\documentclass{article}

\usepackage{amssymb}
\usepackage[figuresright]{rotating}

\usepackage{cite}
\newcommand{\citep}[1]{\cite{#1}}

\usepackage{hyperref}

\usepackage{subcaption}
\usepackage{multirow}
\usepackage{booktabs}
\usepackage{apxproof}
\usepackage{longtable}
\usepackage{algorithm}
\usepackage{algorithmic}

\usepackage{amsmath}
\usepackage{amsfonts}
\usepackage{amssymb}
\usepackage{amsthm}

\usepackage{xypic}

\theoremstyle{definition}
\newtheorem{definition}{Definition}

\theoremstyle{plain}

\theoremstyle{remark}

\newcommand{\N}{\mathbb{N}}
\newcommand{\R}{\mathbb{R}}
\renewcommand{\P}{\mathbb{P}}
\newcommand{\D}{\mathcal{D}}
\renewcommand{\d}{\textnormal{d}}

\renewcommand{\d}{\textnormal{d}}
\renewcommand{\1}{\mathbf{1}}

\newcommand{\indp}{\perp\!\!\!\perp}

\newcommand{\X}{\mathcal{X}}
\newcommand{\T}{\mathcal{T}}

\usepackage{tikz}
\usepackage{pgfplots}
\pgfplotsset{compat=1.8}
\usetikzlibrary{arrows.meta,patterns,decorations.text,fit,calc}
\usepgfplotslibrary{statistics}

\definecolor{tangored}{HTML}{CC0000}
\definecolor{tangoplum}{HTML}{75507B}
\definecolor{tangoblue}{HTML}{3465A4}
\definecolor{tangoorange}{HTML}{F57900}
\definecolor{tangogreen}{HTML}{73D216}
\definecolor{tangobutter}{HTML}{EDD400}

\definecolor{aluminium1}{HTML}{EEEEEC}
\definecolor{aluminium2}{HTML}{D3D7CF}
\definecolor{aluminium3}{HTML}{BABDB6}
\definecolor{aluminium4}{HTML}{888A85}
\definecolor{aluminium5}{HTML}{555753}
\definecolor{aluminium6}{HTML}{2E3436}

\colorlet{class0c}{aluminium4}
\colorlet{class1c}{tangoplum}
\colorlet{class2c}{tangogreen}
\colorlet{class3c}{tangoorange}
\colorlet{class4c}{tangoblue}

\tikzstyle{classinv}=[draw=none, mark size=0pt]
\tikzstyle{class0}=[fill=class0c, draw=aluminium6,mark size=1pt]
\tikzstyle{class1}=[fill=class1c, draw=class1c!80!black,mark size=1pt]
\tikzstyle{class2}=[fill=class2c, draw=class2c!80!black,mark size=1pt]

\pgfmathdeclarefunction{invgauss}{2}{%
	\pgfmathparse{sqrt(-2*ln(#1))*cos(deg(2*pi*#2))}%
}

\pgfdeclarelayer{top}
\pgfsetlayers{main,top}

\newcommand{\labelsize}{\scriptsize}

\pgfplotsset{ticks=none,
	yticklabels=\empty,
	xticklabels=\empty,
	width=0.2\textwidth,
	height=0.2\textwidth,
	label style={font=\labelsize},
}

\newcommand{\plotdataset}[4][1]{
	\pgfmathsetseed{42}
	\addplot [#2,only marks, samples=10] ({invgauss(rnd,rnd)},{invgauss(rnd,rnd)});
	\addplot [#2,only marks, samples=5] ({1.5*invgauss(rnd,rnd)+#4},{invgauss(rnd,rnd)+#1*\yshift});
	\addplot [#3, only marks, samples=10] ({invgauss(rnd,rnd)+2*#4},{invgauss(rnd,rnd)});
	\addplot [#3, only marks, samples=5] ({1.5*invgauss(rnd,rnd)+#4},{invgauss(rnd,rnd)+#1*\yshift});
}

\newcommand{\plotOnlyProtos}{
	\addplot[only marks,class1,mark=diamond*,mark size=3pt](0,\protoshift);			
	\addplot[only marks,class2,mark=diamond*,mark size=3pt](2*\xshift,-\protoshift);
}

\newcommand{\plotSegment}[1][]{
	\pattern[pattern=north east lines, pattern color=class1c] (axis cs:-5,-5)--(axis cs:-5,0.5*\yshift)--(axis cs:\xshift,0.5*\yshift)--(axis cs:\xshift,-5)--cycle;
	\pattern[pattern=north west lines, pattern color=class2c] (axis cs:\xshift,-5)--(axis cs:\xshift,0.5*\yshift)--(axis cs:4*\xshift,0.5*\yshift)--(axis cs:4*\xshift,-5)--cycle;
	\ifthenelse{\equal{#1}{}}{}{
		\pattern[pattern=vertical lines, pattern color=class0c] (axis cs:-5,0.5*\yshift)--(axis cs:-5,2*\yshift)--(axis cs:4*\xshift,2*\yshift)--(axis cs:4*\xshift,0.5*\yshift)--cycle;
	}
}

\def\xshift{3.5}
\def\yshift{8}
\def\protoshift{0.5}

\newcommand{\tikzvoodoo}{
	\begin{tikzpicture}[
		node distance=3em,
		from/.style={{Stealth[length=0.8em, round]}-,double,shorten >=-0.2em,shorten <=-0.2em},
		towards/.style={-{Stealth[length=0.8em, round]},double,shorten >=-0.2em,shorten <=-0.2em},
		]
			\node(beforedrift) {%
				\begin{tikzpicture}
					\begin{axis}[xlabel={before drift}]
						\plotdataset{class1}{classinv}{\xshift}
					\end{axis}
				\end{tikzpicture}
			};
				\node(afterdrift) [node distance=0em, below=of beforedrift] {%	
					\begin{tikzpicture}
						\begin{axis}[xlabel={after drift}]
							\plotdataset{classinv}{class2}{\xshift}
						\end{axis}
					\end{tikzpicture}
				};
				\draw[decorate,decoration={brace,amplitude=1.5em,mirror,raise=-0.3em}] (beforedrift.north west) -- node[xshift=-1.2em](leftdrift){} (afterdrift.south west);
				\draw[decorate,decoration={brace,amplitude=1.5em,raise=-0.3em}] (beforedrift.north east) -- node[xshift=1.2em](rightdrift){} (afterdrift.south east);	
					\node(stream) [below left=of leftdrift] {%
						\begin{tikzpicture}
							\begin{axis}[xlabel={}]
								\plotdataset{class0}{class0}{\xshift}
							\end{axis}
						\end{tikzpicture}
					} edge[towards]
					[postaction={decoration={text along path, text align=center,text={|\labelsize |drift ||},
							raise=0.2em,},decorate}]
					[postaction={decoration={text along path, text align=center,text={|\labelsize |detection ||},
							raise=-0.7em,}, decorate}]
					(leftdrift);
					\node[node distance=0em,below=of stream] {stream};
					\node(driftingarea) [right=of rightdrift] {
						\begin{tikzpicture}
							\begin{axis}
								\plotdataset{class1}{class2}{\xshift}
								\plotSegment
							\end{axis}
						\end{tikzpicture}
					} edge [from] node[above]{\labelsize train} 
					node[below]{\labelsize model} (rightdrift);
					\node[node distance=0em,below=of driftingarea,font=\labelsize,align=center] {drift \\[-0.5em] locus};
					\node (nopreproc) [right=of driftingarea] {
						\begin{tikzpicture}
							\begin{axis}[axis line style={white,}]
								\plotdataset{classinv}{classinv}{\xshift}
							\end{axis}
						\end{tikzpicture}
					} edge [from] (driftingarea);
				    \node at (nopreproc) {\Huge $\bullet$};
					\node[label=east:DiDi] (didi) [right=of nopreproc] {
						\begin{tikzpicture}
							\begin{axis}
								\plotdataset[2]{class1}{class2}{4*\xshift}
							\end{axis}
						\end{tikzpicture}
					} edge [from] (nopreproc);
					\node[label=east:WBM] (wbm) [node distance=0em, above=of didi] {
						\resizebox{.1\textwidth}{!}{
							\begin{tikzpicture}
								\begin{scope}[X/.style={draw,minimum width=3em}]
									\node[X] {$\bullet$}
									child {node[X] {none} edge from parent node[left] {\labelsize $y>0$}}
									child {node[X] {$\bullet$}
										child {node[X] {before} edge from parent node[left] {\labelsize $x\leq0$}}
										child {node[X] {after}edge from parent node[right] {\labelsize $x>0$}}
										edge from parent node[right] {\labelsize $y\leq0$}};
								\end{scope}
							\end{tikzpicture}
						}
					} edge [from] (nopreproc);
					\node[label=east:PFI] (pfi) [node distance=0em,below=of didi] {
							\begin{tikzpicture}
								\begin{axis}[
									boxplot/draw direction=x,
									ymin=0,ymax=5,%ymajorticks=true,
									cycle list={{class1c},{class2c}},
									tick align=inside,
									ytick={1,2,3,4},
									yticklabels={$z_2$,$z_1$,$y$,$x$},
									]
									\addplot+[fill,fill opacity=0.2,
									boxplot prepared={
										median=2,
										upper quartile=2.35,
										lower quartile=1.5,
										upper whisker=3.4,
										lower whisker=1.1
									},
									] coordinates {};
									\addplot+[fill,fill opacity=0.2,
									boxplot prepared={
										median=2.59,
										upper quartile=3.35,
										lower quartile=2,
										upper whisker=4.4,
										lower whisker=1.1
									},
									] coordinates {};
									\addplot+[fill,fill opacity=0.2,
									boxplot prepared={
										median=5.57,
										upper quartile=6.93,
										lower quartile=4.93,
										upper whisker=8.68,
										lower whisker=3.03
									},
									] coordinates {};
									\addplot+[fill,fill opacity=0.2,
									boxplot prepared={
										median=6.57,
										upper quartile=7.93,
										lower quartile=5.93,
										upper whisker=10.68,
										lower whisker=4.03
									},
									] coordinates {};
								\end{axis}
							\end{tikzpicture}
					} edge [from] (nopreproc);
					\node[label=east:CF] (cf) [node distance=0em,below=of pfi] {
						\begin{tikzpicture}
							\begin{axis}
								\plotdataset{class1}{class2}{\xshift}
								\plotSegment
								\plotOnlyProtos
								\begin{pgfonlayer}{top}
									\draw[class1c!80!black,-{Stealth}] (axis cs:0,\protoshift) to [bend left] (axis cs:1.1*\xshift,\protoshift) node[above,midway,black] {};
									\draw[class2c!80!black,-{Stealth}] (axis cs:2*\xshift,-\protoshift) to [bend left] (axis cs:0.8*\xshift,-\protoshift) node[above,midway,black] {};
									\addplot[only marks,class1,mark=triangle*,mark size=2.5pt](1.1*\xshift,\protoshift);			
									\addplot[only marks,class2,mark=triangle*,mark size=2.5pt](0.8*\xshift,-\protoshift);
								\end{pgfonlayer}
							\end{axis}
						\end{tikzpicture}
					};
					\node[label=east:LIME] (lime) [node distance=0em, below=of cf] {
						\begin{tikzpicture}
							\begin{axis}
								\plotdataset{class1}{class2}{\xshift}
								\plotSegment
								\plotOnlyProtos
								\begin{pgfonlayer}{top}
									\draw[class1c!80!black,-{Stealth}] (axis cs:0,\protoshift) -- (axis cs:0.8*\xshift,\protoshift) node[above,near start,black] {\tiny $\partial_1$}; %,yshift=-0.3em
									\draw[class2c!80!black,-{Stealth}] (axis cs:2*\xshift,-\protoshift) -- (axis cs:1.2*\xshift,-\protoshift) node[above,near start,black] {\tiny $\partial_2$}; %midway,yshift=-0.3em
								\end{pgfonlayer}
							\end{axis}
						\end{tikzpicture}
					};% edge [from] (img3b);
					\node[node distance=0em] (mid) at ($(cf.west)!0.5!(lime.west)$) {};
						\node (repproto) at (nopreproc |- mid) {
							\begin{tikzpicture}
								\begin{axis}
									\plotdataset{class1}{class2}{\xshift}
									\plotSegment
									\plotOnlyProtos
								\end{axis}
							\end{tikzpicture}
						} edge [from] (driftingarea);
						\node[node distance=0em,below=of repproto,font=\labelsize,align=center] {representing \\[-0.5em] prototypes};
						\draw (repproto) edge[towards] (cf);
						\draw (repproto) edge[towards] (lime);
						\node (segmentation) [left =of repproto] {
							\begin{tikzpicture}
								\begin{axis}
									\plotdataset{class1}{class2}{\xshift}
									\plotSegment[1]
								\end{axis}
							\end{tikzpicture}
						} edge [towards] (nopreproc);
						\node[node distance=0em,below=of segmentation,font=\labelsize,align=center] {drift \\[-0.5em] segmentation};
						\draw (segmentation) edge [towards] (repproto);
						\node[below right=of stream] {};·
						\node (betweenPfiCf) at ($(pfi.east) !0.5! (cf.east)$) {};
						\node (betweenPfiCfShifted) at ($(betweenPfiCf) + (-20em,0)$) {};
						\draw(stream) edge [towards,out=315,in=180] (segmentation);
						\node[fit=(leftdrift) (beforedrift) (betweenPfiCfShifted) (driftingarea),inner ysep=0em,inner xsep=0.85em] (loc) {}; %
						\node[fit=(segmentation) (loc.south west) (loc.south east),inner ysep=1.7em,inner xsep=0em] (seg) {};
						\draw[dashed,class0c] (loc.north west) -- (seg.south west);
						\draw[dashed,class0c] (loc.south west) -- (betweenPfiCf);%(loc.south east);
						\draw[dashed,class0c] (loc.north east) -- (seg.south east);
						\node[node distance=-0.5em,below=of loc,anchor=south west] (textLoc) {localization};
						\node[node distance=0.5em,below=of loc,anchor=north west] (textSeg) {segmentation};
						\node[anchor=west] at ($(nopreproc.west |- textLoc) + (0.5em,0)$) {global}; %[align=left]
						\node[anchor=west] at ($(nopreproc.west |- textSeg) + (0.5em,0)$) {local};
						
					\end{tikzpicture}
				}

\newcommand{\tikzLocalization}{
\begin{tikzpicture}[
	node distance=3em,
	from/.style={{Stealth[length=0.8em, round]}-,double,shorten >=-0.2em,shorten <=-0.2em},
	towards/.style={-{Stealth[length=0.8em, round]},double,shorten >=-0.2em,shorten <=-0.2em},
	]
		\node(beforedrift) {%
			\begin{tikzpicture}
			\begin{axis}[xlabel={before drift}]
				 \plotdataset{class1}{classinv}{\xshift}
			\end{axis}
			\end{tikzpicture}
		};
		\node(afterdrift) [node distance=0em, below=of beforedrift] {%	
			\begin{tikzpicture}
			\begin{axis}[xlabel={after drift}]
				 \plotdataset{classinv}{class2}{\xshift}
			\end{axis}
			\end{tikzpicture}
		};
	\draw[decorate,decoration={brace,amplitude=1.5em,mirror,raise=-0.3em}] (beforedrift.north west) -- node[xshift=-1.2em](leftdrift){} (afterdrift.south west);
	\draw[decorate,decoration={brace,amplitude=1.5em,raise=-0.3em}] (beforedrift.north east) -- node[xshift=1.2em](rightdrift){} (afterdrift.south east);	
	\node(stream) [left=of leftdrift] {%
		\begin{tikzpicture}
			\begin{axis}[xlabel={}]
				 \plotdataset{class0}{class0}{\xshift}
			\end{axis}
		\end{tikzpicture}
	} edge[towards]
		[postaction={decoration={text along path, text align=center,text={|\labelsize |drift ||},
			raise=0.2em,},decorate}]
		[postaction={decoration={text along path, text align=center,text={|\labelsize |detection ||},
		raise=-0.7em,}, decorate}]
 (leftdrift);
 	\node[node distance=0em,below=of stream] {stream};
	\node(driftingarea) [right=of rightdrift] {
		\begin{tikzpicture}
		\begin{axis}
			\plotdataset{class1}{class2}{\xshift}
			\plotSegment
		\end{axis}
	\end{tikzpicture}
	} edge [from] node[above]{\labelsize train} 
	node[below]{\labelsize model} (rightdrift);
	\node[node distance=0em,below=of driftingarea,font=\labelsize,align=center] {drift locus = region with \\[-0.5em] high class certainty};
	\end{tikzpicture}
 }
\usetikzlibrary{positioning,shapes,arrows}

\newcommand{\bayesTicks}{
\begin{tikzpicture}[
  node distance=1cm and 0cm,
  mynode/.style={draw,ellipse,text width=2cm,align=center},
  Tnode/.style={text width=0.75cm,align=center},
  Inode/.style={draw,very thick,circle,text width=0.5cm,align=center},
  Fnode/.style={draw,very thick,dashed,circle,text width=0.5cm,align=center},
  Nnode/.style={draw,circle,text width=0.5cm,align=center}
]
\node[Fnode                   ] (N01) {$F_1$}; 
\node[Inode,below right=of N01] (N04) {$I_4$}; 
\node[Fnode,above right=of N04] (N02) {$F_2$}; 
\node[Inode,below      =of N04] (N08) {$I_2$};
\node[Fnode,below left =of N08] (N10) {$F_5$}; 
\node[Inode,above left =of N10] (N07) {$I_1$}; 
\node[Inode,above      =of N07] (N03) {$I_3$}; 
\node[Tnode,above right=of N07] (N05) {$T$}; 
\node[Tnode,above right=of N08] (N06) {$T$}; 
\node[Fnode,below left =of N07] (N09) {$F_4$};  
\node[Fnode,below left =of N09] (N13) {$F_6$}; 
\node[Fnode,below right=of N09] (N14) {$F_7$}; 
\node[Fnode,above left =of N09] (N16) {$F_3$};

\node[Nnode,above      =of N16] (N15) {$N_3$}; 
\node[Nnode,above left =of N15] (N11) {$N_1$}; 
\node[Nnode,above right=of N15] (N12) {$N_2$};

\path (N01) edge[-latex] (N04);
\path (N02) edge[-latex] (N04);
\path (N03) edge[-latex] (N07);
\path (N04) edge[-latex] (N08);
\path (N05) edge[-latex] (N07);
\path (N06) edge[-latex] (N08);
\path (N07) edge[-latex] (N09);
\path (N07) edge[-latex] (N10);
\path (N08) edge[-latex] (N10);
\path (N09) edge[-latex] (N13);
\path (N09) edge[-latex] (N14);
\path (N11) edge[-latex] (N15);
\path (N12) edge[-latex] (N15);
\path (N16) edge[-latex] (N09);

\path (N11) edge[-latex] (N12);
\path (N14) edge[-latex] (N13);
\path (N01) edge[-latex] (N02);

\end{tikzpicture}
}

\usepackage{lscape}

\begin{document}

\title{Model based Explanations of Concept Drift}

\author{	
Fabian Hinder,
Valerie Vaquet, \\
Johannes Brinkrolf, and
Barbara Hammer \\\;\\
\textbf{\small\texttt{\{
\href{mailto:fhinder@techfak.uni-bielefeld.de}{fhinder}, 
\href{mailto:vvaquet@techfak.uni-bielefeld.de}{vvaquet}, \href{mailto:jbrinkro@techfak.uni-bielefeld.de}{jbrinkro},  \href{mailto:bhammer@techfak.uni-bielefeld.de}{bhammer}\}@techfak.uni-bielefeld.de}} \\\;\\
\small Bielefeld University - Cognitive Interaction Technology (CITEC)  \\
\small Inspiration 1, 33619 Bielefeld - Germany}
\maketitle

\begin{abstract}
The notion of concept drift refers to the phenomenon that the distribution generating the observed data changes over time. If drift is present, machine learning models can become inaccurate and need adjustment.
While there do exist methods to detect concept drift or to adjust models in the presence of observed drift, the question of \emph{explaining} drift, i.e., describing the potentially complex and high dimensional change of distribution in a human-understandable fashion, has hardly been considered so far. 
This problem is of importance since it enables an inspection of the most prominent characteristics of how and where drift manifests itself. Hence, it enables human understanding of the change and it increases acceptance of life-long learning models.
In this paper, we present a novel technology characterizing concept drift in terms of the characteristic change of spatial features based on various explanation techniques. 
To do so, we propose a methodology to reduce the explanation of concept drift to an explanation of models that are trained in a suitable way extracting relevant information regarding the drift. This way a large variety of explanation schemes is available. Thus, a suitable method can be selected for the problem of drift explanation at hand. 
We outline the potential of this approach and 
 demonstrate its usefulness in several examples.

\end{abstract}

\section{Introduction}
\label{sec:intro}
The world that surrounds us is undergoing continuous changes, which inflict themselves on the increasing amount of available data sources. Those changes occur, for example, in social media or IoT devices, where data is collected over time \cite{DBLP:journals/adt/BifetG20,DBLP:journals/widm/TabassumPFG18a}.
Such effects -- referred to as concept drift -- can be induced by several causes, e.g., seasonal changes, changed demands of individual customers, aging of sensors, etc. %
When dealing with such non-stationary environments, there are two main problem setups: Autonomously running systems need to robustly solve a given task in the presence of drift, and monitoring systems need to reliably detect anomalous behavior. 

The majority of approaches of the first category aim at developing models which adapt in the presence of drift. Usually, the main objective is minimizing the interleaved test-train error. In case autonomous adaption mechanisms fail and human intervention is required, the user is interested in an explanation for potential drops of accuracy~\citep{DBLP:journals/cim/DitzlerRAP15}.

In system monitoring and in drift detection for so-called active methods in non-stationary environments
the drift itself is of interest as it might indicate that certain actions have to be taken~\cite{Aminikhanghahi:2017:SMT:3086013.3086037,DBLP:journals/kais/GoldenbergW19}. For example, in cyber-security settings, drift indicates a potential attack and in the monitoring of critical infrastructure, e.g. electrical grids or water distribution networks, leakages or other failures can cause drift in the observed data~\citep{haes2019survey,eliades_fault_2010}. 
In such cases, more precise information about the drift, ideally some kind of intuitive explanation, can help to minimize the damage caused by a malfunctioning technical system and reduce the wastage of resources by providing more information to the human operator.

In recent years, considerable research has been conducted on explainable AI. Explanation technologies for classical batch machine learning models 
can be stratified according to different questions, ranging from  how an explanation is computed in relation to the given model (e.g.\ black-box, post-hoc, natively explainable), what is explained (e.g.\ local or global behavior), to  which explanations are used (e.g.\ feature-based or example-based) \cite{10.1145/3236009}.
Just as there are several types of machine learning models each with its own strengths, weaknesses, and use cases, there is also a large variety of explanation schemes~\cite{Ribeiro2016WhySI,DeepView,venna2010information,breiman2001random,nilsson2007consistent,shapley1951notes,fumagalli2022incremental,simonyan2013deep,CounterfactualWachter,looveren2019interpretable}, with each putting focus on a different objective and providing different information for the problem at hand.

Research on analyzing and explaining drift is still limited. Methods for an inspection of the most significant aspects. They can explain the drift, currently mostly focusing on comparably narrow aspects, which are usually not sufficient if a precise description of the drift is required, as is typically the case for high dimensional data. They address the questions when the drift occurs (drift detection) and analyze its strength (drift quantification) \cite{Lu_2018}. First technologies aim for the identification of particularly relevant features, i.e. those where drift occurs \cite{DBLP:journals/corr/WebbLPG17,263854}. These approaches are accompanied by methods that attempt identification of inconsistencies caused by drift, i.e. identification of those parts of the model which need to be exchanged, or drift localization, i.e., the parts of the data distribution that are undergoing drift \cite{Lu_2018}. Although they can theoretically take more complex forms of drift into account, they usually do not provide a condensed explanation. However, such are necessary for human users in general, for pure domain experts as an accessible explanation and even more so for the general audience.

The purpose of our contribution is to provide a novel formalization of how to explain observed drift %
such that informed monitoring of the underlying process and the characteristics of the change becomes possible, even for high dimensional, non-sematic data, like images.
More specifically, we show how to make use of the link between the problems of drift localization~\cite{Lu_2018,localization} and drift segmentation~\cite{segmentation} which can be performed by usual probabilistic classification or conditional density estimators, to obtain an efficient algorithmic scheme for drift  explanations using various schemes of model explanations. As such characteristics can be %
complex, they allow for a detailed %
inspection of the drift while still being understandable for the human user and sufficiently rich to %
describe the specific problem at hand.

This paper is organized as follows: 
In Section~\ref{sec:setup} we recall the formal definition of concept drift (Section~\ref{sec:setup:framework}), give a comparative overview of the existing literature for drift explanations (Section~\ref{sec:setup:relatedwork}), and provide a high level description of some of the most relevant explanation schemes (Section~\ref{sec:setup:xai}) as well as drift localization and segmentation (Section~\ref{sec:setup:localization}). %
In Section~\ref{sec:explain} we outline the proposed technology in general and then focus on specific problems and instantiations. Then we illustrate the proposed method in Section~\ref{sec:experiments} using several examples, starting from a more quantitative evaluation and proceeding to more and more complex problem setups. Finally, we conclude and point out some further research questions and problems (Section~\ref{sec:conclusion}).

\section{Problem Setup and Related Work}
\label{sec:setup}
In this work, we propose a methodology for drift explanation relying on model-based approaches which extract spatial properties of drift \cite{segmentation,localization} and several methods for model explanations~\cite{Ribeiro2016WhySI,DeepView,venna2010information,breiman2001random,nilsson2007consistent,shapley1951notes,fumagalli2022incremental,simonyan2013deep,CounterfactualWachter,looveren2019interpretable}.
Before recapping these, we recall the formal framework for concept drift as introduced in the work \cite{DAWIDD,ContTime} and provide a summary of the related work on explanation methods in the context of concept drift.

\subsection{A Statistical Framework for Concept Drift}
\label{sec:setup:framework}
In the classical batch setup of machine learning one considers a generative process $\D$, i.e.,\ a probability measure, on a data space $\X$. In this context one views the realizations of i.i.d.\ random variables $X_1,...,X_n \sim \D$ as samples.
However, this setup is not applicable in many real-world applications where the data is arriving consecutively over time. Thus, it is necessary to adapt the problem description:  
One considers an index set $\T$, representing time, and a collection of (possibly different) distributions $\D_t$ on $\X$, indexed over $\T$~\cite{asurveyonconceptdriftadaption}. It is possible to extend this setup to a general statistical interdependence of data and time via a distribution $\D$ on $\T \times \X$ which decomposes into a distribution $\P_T$ on $\T$ and the conditional distributions $\D_t$ on $\X$ such that $X \mid T = t \sim \D_t$~\cite{DAWIDD,ContTime}.

Drift refers to the fact that $\D_t$ varies for different timepoints, i.e.\ 
$\{ (t_0,t_1) \in \T^2 : \D_{t_0} \neq \D_{t_1} \}$
has measure larger zero w.r.t\ $\P_T^2$~\cite{DAWIDD,ContTime}. 
One of the key findings of \cite{DAWIDD,ContTime} is a unique characterization of the presence of drift by the property of  statistical dependency of time $T$ and data $X$ if a time-enriched representation of the data $(T, X) \sim \D$ is considered.

\subsection{Related Work}
\label{sec:setup:relatedwork}
While explainability has been a major research interest in recent years \cite{molnar2019,DBLP:journals/tamd/RohlfingCSMBBEG21}, explainability methods for drift are still limited.
Quite a number of approaches aim for the detection and quantification of drift  \cite{Lu_2018,AnalyzingConceptDriftAndShiftFromSampleData}, its localization in space \cite{Lu_2018}, or visualization \cite{wang2020conceptexplorer,DBLP:journals/corr/WebbLPG17,DBLP:journals/corr/WebbLPG17,10.1145/956750.956849}.
Besides, several methods focus on feature-wise representations of drift \cite{wang2020conceptexplorer,AnalyzingConceptDriftAndShiftFromSampleData,DBLP:journals/corr/WebbLPG17,10.1145/956750.956849}.
However, they are limited if dealing with high-dimensional data or non-semantic features. 
An exception is \cite{263854} which takes large-scale feature correlation into account by applying a LIME~\cite{Ribeiro2016WhySI} like procedure to contrastive autoencoders. %
To our knowledge, this is the only other approach relying on more complex XAI methods for explaining drift.
However, this approach is limited to the (semi-)supervised setup and provides only relevant features, which might be less intuitive to humans \cite{ijcai2019-876}.
To explain drift, more information than that is required to select change points or estimate the rate of change must be extracted. To detect drift, a single drifting feature is sufficient; to explain drift, all are desirable. 
Drift localization offers such information~\cite{localization} but has mostly been proposed as a subroutine of drift detection rather than explicit drift explanation technology so far: 
All methods \cite{ijcai2017-317,LSDD,locationofdrift,DBLP:journals/ai/LuLZM16} summarized in the overview \cite{Lu_2018} are restricted to a measurement  of the  local change of the distribution rather than an explanation by means of XAI techniques, which can track reasonable directions of drift over time.
Our proposed method relies on ideas as introduced in the work \cite{esann2022,localization,segmentation} as a subroutine to extract information regarding the change. We will explain those in more detail in Section~\ref{sec:setup:localization}. 

\subsection{General XAI for Model Explanations}
\label{sec:setup:xai}
There are several approaches to explainable AI (XAI). Generally, explanation methods can be categorized with respect to the way information is presented: global explanations describe the model as a whole while local or sample-based explanations provide an explanation for a single sample and how the model processed it. In Section~\ref{sec:explain:method} we will show how to extract particular relevant information about the drift using machine learning models and thereby link the problem of general model explanations and explanation of drift. This way we obtain a very broad range of explanation methods which can be fitted to the specific problem at hand. 

In the following, we will recall some of the most relevant approaches and discuss their strengths and weaknesses. 
In Section~\ref{sec:explain:combinations}, we will discuss how the methods listed below can be applied and interpreted in the drift-specific setup and also provide an overview regarding the method-specific characteristics (see Table~\ref{tab:explain:combinations:overview}). Showcases for all explanation schemes, except for interpretable models which we exclude due to their limited explanatory complexity, are provided in Section~\ref{sec:experiments}.

\paragraph*{Interpretable Models} One of the simplest approaches to explainable AI are white-box or interpretable models~\cite{molnar2019,du2019techniques} which by design allow analysis by the user. Typical models of this category are linear models, decision trees, and prototype-based models. However, whether or not a specific model is actually interpretable depends on the concrete setup. For example on image data, neither linear models nor decision trees are intuitively interpretable due to a large number of features. Complex decision trees with several hundred leaves also lack interpretability. %
This is due to the fact that the amount of information a human can take in at once is limited. Thus, the complexity of interpretable models -- or all global explanations for that sake -- has to be rather limited. This in turn limits the complexity of the models and thus the number of applications for this approach. 

\paragraph*{Discriminative Dimensionality Reduction} This group of techniques constitutes another global explanation approach. %
Standard dimensionality reduction approaches are refined with additional information obtained from a machine learning model. 
For example in \cite{venna2010information,DeepView}, the used metric is enriched with information on the decision boundary of the model so that the global decision structure becomes accessible to the user. In contrast to interpretable models, there are no problems with model complexity in this approach. However, it leads to a possible loss of information during the dimensionality reduction process. Thus, for more complex domains or problems, this approach might simplify too much in order to obtain a full picture of the problem at hand.

\paragraph*{Global Feature Importance} One of the oldest inspection methods is permutation feature importance~\cite{breiman2001random}.
The basic idea is to obtain knowledge on the relevance of the single features for the model's internal decision process by permuting the values of a single feature and comparing the model's performance on the modified dataset to the performance on the original dataset.
Comparable approaches that are more theoretically grounded are provided by feature relevance theory~\cite{nilsson2007consistent}, which is linking conditional independence and graphical models, or Shapley-Values~\cite{shapley1951notes}, which are resulting from game-theoretic modeling of feature importance from a set of axioms. %
The drawback of this approach is that it does not provide any information about the single sample as the information is presented in a cumulative fashion only. 

In this work, we will mainly focus on permutation feature importance as a very efficient and model-agnostic approach. This method has the benefit that a new variation allows computing those values in an incremental fashion~\cite{fumagalli2022incremental} which is of particular relevance in the considered online setup.

\paragraph*{Local Feature Importance} A counterpart to the global feature importance methods is provided by Saliency Maps~\cite{simonyan2013deep} or Local Interpretable Model-agnostic Explanations (LIME)~\cite{Ribeiro2016WhySI}. 

Saliency Maps originate from explainable AI in deep learning, in particular image classification. The idea is to find the most important features of a sample by computing the gradient of the classification function with respect to the sample in question. The features associated with the absolute largest partial derivative are then considered as particularly relevant as a change of those would result in a particularly large change of the classification. One obvious drawback of this method is that derivatives are subject to local disturbances of the classification function. This poses a problem as those might be an artifact of the model and do not have any implication on the actual classification.

LIME on the other hand aims to provide an explanation for the classification of a single sample by means of a simplified model, which can be easily interpreted by the user: a model is trained on the complex model's predictions on samples in the proximity of the sample in question. Linear models constitute a common choice for the simplified model. In this case, LIME essentially computes the derivative of the convolution of the model and the sampling distribution. Therefore, LIME suffers -- at least to some extent -- from the same issues as Saliency Maps. The sampling distribution is a crucial parameter of the method.

\paragraph*{Contrasting Explanations and Counterfactuals}
One way to tackle the problem of local disturbances is provided by counterfactual explanations~\cite{CounterfactualWachter}. While derivatives only provide a hint on how a sample has to be changed in order to obtain a different class counterfactual explanations actually provide a sample of the other class, called a counterfactual. %
In order to allow the user to grasp the most relevant changes or characteristics of the decision process for the sample at hand, counterfactuals should have additional properties such as being as close as possible to the original sample (closest counterfactual) or lying on the data manifold (plausible counterfactual)~\cite{molnar2019,looveren2019interpretable}.
Drawbacks of this method are the computational cost, missing uniqueness of the obtained counterfactuals, and their conceptual closeness to adversarial examples.

\subsection{Spatial Properties of Concept Drift: Drift Localization and Drift Segmentation}
\label{sec:setup:localization}

\begin{figure}[!t]
    \centering
    \begin{align*}
        \xymatrix{
        & h \ar@{|->}[rd]^{\text{explain model}}& \\
        \D_t \ar@{|->}[ru]^{\text{train model}} \ar@{|..>}[rr]_{\text{explain drift}} & & \mathcal{E}
        }
    \end{align*}
    \caption{General Scheme for drift explanation: 1. train a model ($h$) to capture the relevant information of the drift ($\D_t$), 2. extract information from the model ($h$) via model  explanations to obtain an explanation ($\mathcal{E}$) for the drift.}
    \label{fig:generalscheme}
\end{figure}
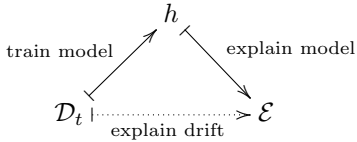

As stated above the main idea of this contribution is to apply model based explanations to concept drift: %
First use a model to learn relevant characteristics of the drift and then compute the explanation of the model to understand the drift by proxy (see Figure~\ref{fig:generalscheme}). The tasks which we will consider, aim at an extraction of relevant spatial properties of the drift: Drift Localization and Drift Segmentation. In Section~\ref{sec:explain:method} we will discuss that such tasks already suffice to obtain and therefore explain all relevant information of the drift. In this section, we recall the general tasks of drift localization and segmentation and elaborate on how they are connected to and can be reduced to a common learning problem.  

\paragraph*{Drift Localization} The problem of identifying whether or not a specific sample is affected by drift or equivalently of finding the regions in dataspace where the drift manifests itself is referred to as drift localization. We illustrated this idea in Figure~\ref{fig:setup:localization:localization}. The task usually assumes a finite collection of timepoints, i.e., $|\T| < \infty$. 
In the following, we will assume that drift has been detected. %
This allows us to segment the data stream into the time before and after the drift, corresponding to two timepoints $\T = \{0,1\}$.
Localizing the drift can then be defined as finding a minimal set $L \subset \X$ such that the distributions coincide for the rest: $\D_0(A \setminus L) = \D_1(A \setminus L)$ for all measurable sets $A$:

\begin{figure}[!t]
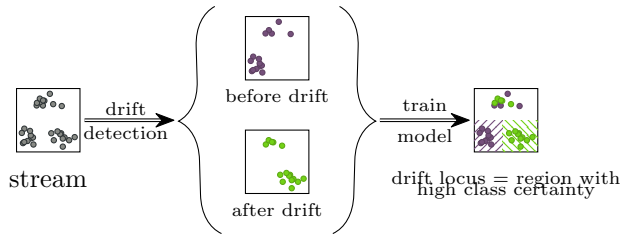

    \centering
    \tikzLocalization
    \caption{Schematic visualization of drift localization. First, by means of drift detection obtain labeling before/after drift. In a second step, train a model. Regions of high certainty correspond to the drift locus.}
    \label{fig:setup:localization:localization}
\end{figure}

\begin{definition}[(Minimal) Drift Locus, Drift Localization~\cite{localization}]
\label{def:localization}
Let $(\D_t,\P_T)$ be a drift process~\cite{DAWIDD}. 
This induces the measure $\D_\T(A) := \int \D_t(A) \d \P_T(t)$ as the mean distribution of $X$ over time.
A \emph{drift locus} is a measurable set $L \subset \X$ such that $(\D_t( \cdot |L^C),\P_T)$ has no drift and $t \mapsto \D_t(L)$ is $\P_T$-a.s. constant. A drift locus $L$ is \emph{minimal} if it is contained in every other drift locus $L'$ up to a $\D_\T$-null set, i.e., $\D_\T(L \setminus L')= 0$.
We refer to the process of finding the minimal drift locus as \emph{drift localization}.
\end{definition}

The minimal drift locus $L$ is is uniquely determined by this property.
In particular, it can be shown that $L$ is not empty if and only if there is drift~\cite[Lemma 1]{localization}.%

Interestingly, drift localization can be realized by training a probabilistic classification model $h : \X \to \Pr(\T)$ to predict the time $t$ given a sample $x$, i.e., whether $x$ was observed before ($t = 0$) or after ($t = 1$) the drift~\cite[Theorem 2 and Corollary 2]{localization}: If the probability coincides with the prior probability up to statistical insufficiencies, i.e., $h(x) \approx \P_T$, then there is no drift and vice versa (see Algorithm~\ref{alg:setup:localization:localization}). Thus, the model that we have to explain is the classifier $h$. 

In the special case of $\T = \{0,1\}$ we can further refine the drift locus into those parts which are more likely to occur before and those that occur after the drift which we will refer to as \emph{drift regions}.

Notice that by explaining $h$ we can obtain even more fine-grained explanations of the drift than provided by the localization itself as it actually measures the local drift intensity, which is continuous, rather than just the binary drift locus. However, one major drawback of drift localization is that it can only be applied if we consider two timepoints, i.e., ``before'' and ``after'' the drift. This requires drift detection, which itself is a non-trivial task. 

\begin{algorithm}[t]
   \caption{Drift localization.}
   \label{alg:setup:localization:localization}
\begin{algorithmic}[1]
   \STATE {\bfseries Input:} {$S = \{(x_1,t_1),...,(x_n,t_n)\}$ dated datapoints, $\theta$ decision threshold}
   \STATE {\bfseries Output:} {$L$ drift locus at datapoints}
   \STATE $t_\text{drift} \gets \textsc{DriftDetection}(S)$ \label{alg:setup:localization:localization:preprocess}\COMMENT{Determine change point}
   \STATE $S' \gets \{ (x_i,t_i') \mid i = 1,\dots,n \} \text{ with } t_i' = \1[t_i \geq t_\text{drift}]$
   \STATE $L \gets \textsc{Zeros}(n)$
   \FORALL{$k = 1,...,n$}
   \STATE $h \gets \textsc{TrainProbabilisticModel}(\{(x_i,t_i') \in S' \mid i \neq k\})$
   \STATE $h_0 \gets \frac{1}{n-1} \sum_{i \neq k} t_i' $\label{alg:setup:localization:localization:train}
   \STATE $L[k] \gets \1[ D_{\text{KL}}(h(x_k)\Vert h_0) \geq \theta ]$
   \ENDFOR
   \RETURN $L$
\end{algorithmic}
\end{algorithm}

\paragraph*{Drift Segmentation} The task of subdividing the dataspace $\X$ into regions or segments of homogeneous drift behavior is referred to as drift segmentation. As discussed in \cite{segmentation} the drifting behavior at the point $x \in \X$ is encoded in the conditional distribution $\P_{T \mid X = x}$. Thus, formally we want to obtain a map $L : \X \to \N$ such that $L(x) = L(x')$ implies $\P_{T\mid X=x} = \P_{T \mid X=x'}$: 

\begin{definition}[Drift Segmentation \cite{segmentation}]
For a drift process $(\D_t,\P_T)$ with $(X,T) \sim \D$ a \emph{drift segmentation} of is a measurable map $L : \X \to \N$, 
 which assigns each element of $x \in \X$ an index $L(x)$ corresponding to the segment it belongs to, such that
 $\P_{T|L(X)}=\P_{T|X}$. 
\end{definition}

It can be shown that $L$ is a drift segmentation if and only if $T \indp X \mid L(X)$~\cite[Lemma 1]{segmentation} which essentially means that if one considers one of the segments $L^{-1}(i),\;i \in \N$ only one does not observe any drift but only a fluctuation of occurrence probability. 

In some sense drift segmentation is a continuous extension of drift localization. Indeed, as already pointed out by \cite{segmentation} one can turn any drift segmentation into a drift localization by marking all segments with $\P_{T \mid L(X)} \neq \P_{T}$ as drifting. However, besides that, it does not require a finite set of timepoints, and thus drift detection as a prepossessing. Furthermore, drift segmentation is also more fine-grained compared to drift localization. This makes it better suited for continuous time as continuous drift may lead to a situation where the entire dataspace is undergoing drift, although the local drift characteristics might vary a lot.

\begin{algorithm}[t]
   \caption{Drift segmentation.}
   \label{alg:setup:localization:segmentation}
\begin{algorithmic}[1]
   \STATE {\bfseries Input:} {$S = \{(x_1,t_1),...,(x_n,t_n)\}$ dated datapoints, $d$ degree of preprocessing}
   \STATE {\bfseries Output:} {$L$ drift segment at datapoints}
   \STATE $S' \gets \{(x_i,(t_i,t_i^2,\dots,t_i^d))\}$\label{alg:setup:localization:segmentation:preprocess}
   \STATE $h \gets \textsc{TrainSegmentationBasedMultiRegressionModel}(S')$\label{alg:setup:localization:segmentation:train}
   \STATE $l \gets \textsc{ExtractSegmentationFunction}(h)$ \COMMENT{$h(x) = v \circ l(x)$ with $l : \X \to \N$, $v: \N \to \R^d$}
   \STATE $L \gets \textsc{Zeros}(n)$
   \FORALL{$k = 1,...,n$}
   \STATE $L[k] \gets l(x_k)$
   \ENDFOR
   \RETURN $L$
\end{algorithmic}
\end{algorithm}

To obtain a drift segmentation \cite{segmentation} suggested training a special type of decision trees called Kolmogorov-trees, which are trained using the Kolmogorov-Smirnov test, and consider the leaves of those trees as drift segments. Following the ideas of \cite{momentTrees} one can also obtain a drift segmentation by training a classical decision tree for the multi-regression problem $X \mapsto (T,T^2,\dots,T^d)$:
By \cite[Lemma 1]{segmentation} a map $L : \X \to \N$, which can be given by a decision tree, is a drift segmentation if $T$ and $X$ are independent given $L(X)$. Since $L$ is deterministic, we can measure conditional independence by 
$\Vert \P_{T \mid X} - \P_{T \mid L(X)} \Vert_W$. Using \cite[Theorem 2]{momentTrees} this can be upper bounded by the variance of $(T,T^2,\dots,T^d)$ given $L(X)$, i.e., the MSE of the obtained decision tree, and a constant term that depends on $d$ and goes to zero for $d \to \infty$.

This once again shows the close connection to drift localization as a decision tree trained for the case $\T = \{0,1\}, d = 1$ learns the conditional class probability of $T = 1$, i.e., is a probabilistic classifier.

Notice that decision trees are not the only method to obtain a drift segmentation. Indeed, using the ideas from \cite{momentTrees,izbicki2017converting} every segmentation-based, multi-regression model gives rise to a drift segmentation (see Algorithm~\ref{alg:setup:localization:segmentation}). %

\section{Explaining Concept Drift}
\label{sec:explain}

When monitoring processes, it is crucial to detect anomalous behavior. Drift detection technologies can automate this step and identify the point in time where the distribution changes \cite{eddm,adwin,LSDD,hdddm,ddm,pagehinkley,Wald}. However, knowing only the time of the drift is usually not sufficient as it often remains unclear how to react to such drift, i.e.,\ to decide whether adapting the model, redoing an analysis, or human intervention is required. While the detected anomaly might be analyzed by an expert, it would be more efficient to obtain explanations in an automated fashion. This would enable a human to initiate an appropriate reaction or, for a layperson, increase an understanding of the necessity to update the model.
A drift characterization is particularly demanding for high dimensional data or a lack of clear semantic features. 
However, even in cases with semantic features capturing the drift, i.e., finding the features affected by the drift, can be challenging in its own right. In particular, this is the case if not the features themselves but only their correlation is affected by the drift. 

Although this gives rise to a seemingly large number of problems, many of them can be tackled using a general framework which we are going to present in the following. We will start by outlining the core ideas with an illustrative example. We will then continue by discussing the general algorithm and methodology followed by highlighting some more specific instantiations of this idea. 
\subsection{Outline of Proposed Method(s): An Example}
\label{sec:explain:example}
In this example, we rely on an example-based explanation scheme. We propose to describe the drift characteristics by contrasting suitable representatives of the underlying distributions \cite{CounterfactualWachter,molnar2019}. 
Before delving into more details, let us describe the underlying motivation: 

\begin{figure}[!t]
    \centering
    \begin{minipage}{0.3\textwidth}
    \centering
    \includegraphics[width=0.9\textwidth]{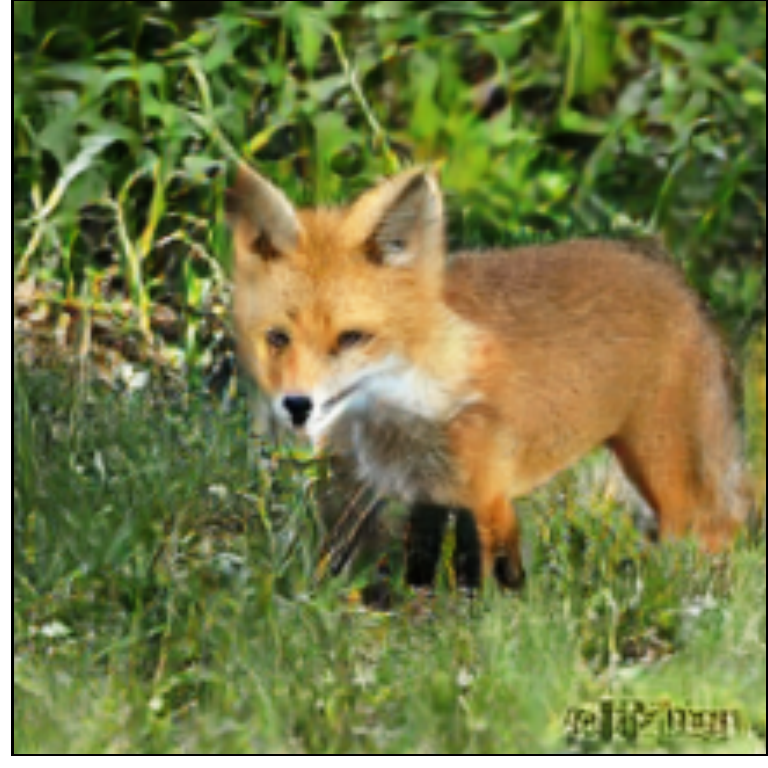}
    \subcaption{Prototype}
    \end{minipage}
    \begin{minipage}{0.1\textwidth}
    \centering
    \Large
    $\Rightarrow$
    \end{minipage}
    \begin{minipage}{0.3\textwidth}
    \centering
    \includegraphics[width=0.9\textwidth]{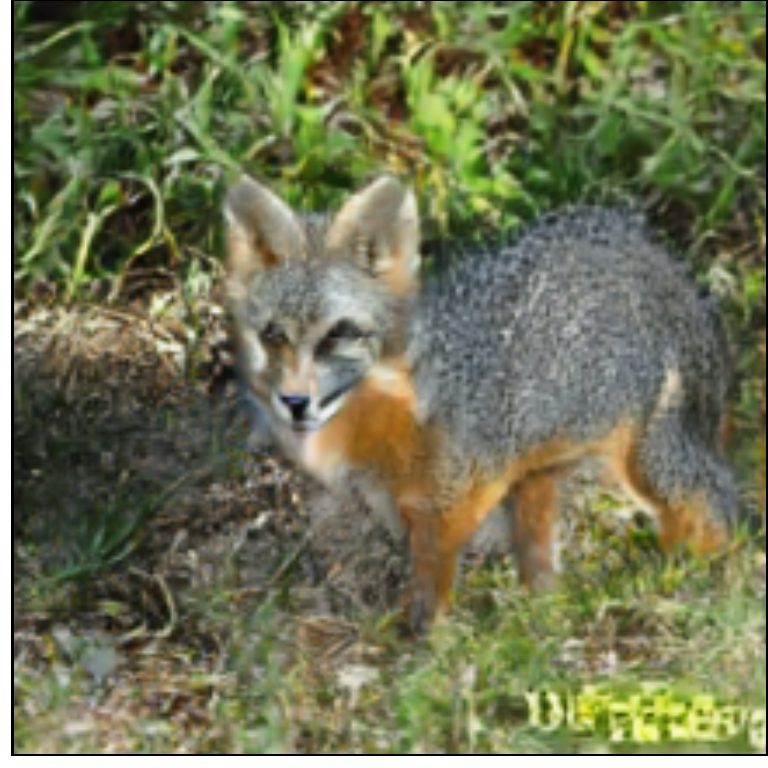}
    \subcaption{Counterfactual}
    \end{minipage}
    \caption{Illustration of a representative example for the before drift regions (red foxes will vanish) and associated counterfactual (gray foxes ``replace'' the red ones).}
    \label{fig:zooexamplepicture}
\end{figure}

Suppose we are considering steams of pictures taken by stationary webcams in a zoo. If the species within a compound are exchanged, say red fox by gray fox, it will cause drift. %
Assume that we already applied drift detection and obtained the time of the drift, i.e., we are dealing with determined cases ``before'' and ``after''. Explaining the drift to a user in an intuitive way can be achieved by presenting two pictures showing the compound before and after the drift, as shown in Figure~\ref{fig:zooexamplepicture}. Inspecting these images, the user can easily spot the difference and understand that the fox species changed.

In this example, the species can be considered as a feature, that appears or ceases to be. Geometrically speaking, the distribution moves between the regions where the feature is or is not present. Thus, the change can be explained by localizing the drift and representing the localization in an intuitive way. Drift localization can essentially be performed by first training a classifier discriminating between samples collected before and after the drift occurred and then analyzing it (we will discuss the validity of this approach in Section~\ref{sec:explain:method}). 
The intuition behind this idea is rather simple: Algorithmically we can find characteristics, just as the species in the example above, using machine learning models. If a sample shows a certain feature -- for example the species -- that did only occur at a certain point in time, that sample cannot be observed at any other date, and can therefore be used to predict the moment of observation.
Finally, a suitable explanation needs to be computed and presented to the user. In this example, we rely on contrastive explanations, since they are considered particularly easy to understand by humans. As the species is the main feature to date the picture, the system would try to produce a counterfactual explanation consisting of the original picture and a counterfactual where the red fox was retouched by a gray fox, allowing the user to grasp the feature ``species'' as desired.

In order to perform this task in a completely automated fashion we also need to find samples where the drift manifests itself, i.e., show those features in a particularly prominent way. For this, we will rely on weighted, prototype-based clustering methods (see Section~\ref{sec:explain:characteristicsamples}). 

By applying this approach to a data stream that simulates our zoo example using ImageNet~\cite{imagenet} pictures, we obtain the explanation given in Figure~\ref{fig:zooexamplepicture} which we discuss in more detail in Section~\ref{exp:imagenet}.

\begin{figure}[!t]
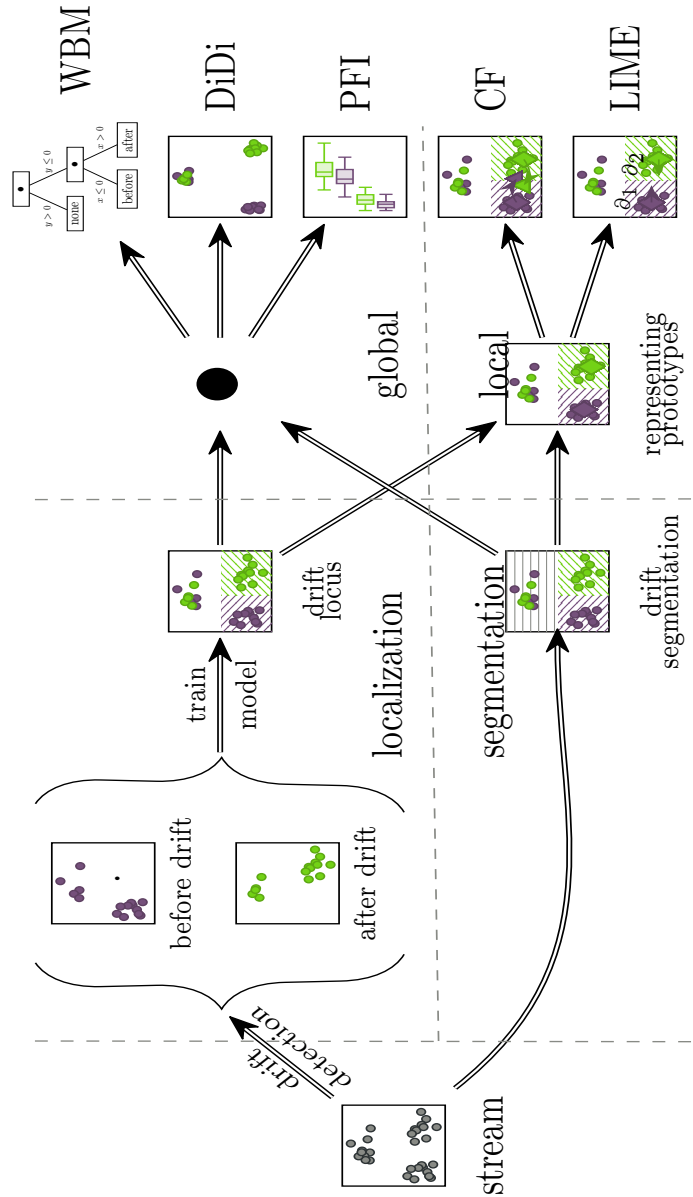

    \centering
    \resizebox{0.8\textwidth}{0.85\textheight}{
    	\rotatebox{90}{\tikzvoodoo}
    }
    \caption{Outline of solutions and processing steps. The stream is first partitioned by means of drift localization or drift segmentation. In a second step, global or local explanation techniques are applied. For local explanations first representing prototypes need to be obtained. 
    }
    \label{fig:explain:combinations:outlines}
\end{figure}

\subsection{General Algorithm and Methodology}
\label{sec:explain:method}
The core idea for our explanation approach is essentially based on an application of the Bayes Theorem. If $\D_t$ is a drift process and $X,T$ are data time pairs drawn from it, i.e., $(X, T) \sim \D$. Then by Bayes Theorem, for time windows $W \subset \T$ and areas in dataspace $A \subset \X$ it holds
\begin{align*}
    \D_W(A) = \P[X \in A \mid T \in W] = \frac{\P[X \in A]}{\P[T \in W]} \P[T \in W \mid X \in A]. 
\end{align*}
As pointed out by \cite{DAWIDD}, drift is encoded in the dependence of $X$ and $T$. Thus, the term ${\P[X \in A]}/{\P[T \in W]}$ does not contain any information regarding the drift as it does not contain any information regarding the joint distribution of $X$ and $T$. Hence, the entire information is encoded in $\P[T \in W \mid X \in A]$. As $\T$ is usually a subset of the real or natural number the estimation of $\P[T \mid X]$ is essentially a conditional density estimation or probabilistic classification if $\T$ is finite, respectively. Thus, by training a model $h(t \mid x)$ to estimate $\P[T = t \mid X = x]$ we essentially extract information about the drift. 

Indeed, both drift localization and segmentation as approached in \cite{localization,segmentation} and described in Algorithm~\ref{alg:setup:localization:localization} and~\ref{alg:setup:localization:segmentation} perform the task by analyzing such a model for a fix $x$. This directly shows why those approaches are local in the dataspace and that they are strongly connected to the presence of drift, as in the case of the absence of drift, $h(t \mid x)$ becomes $x$-invariant.

On an algorithmic level, the problem of drift explanation can be solved by first training a machine learning model to either localize or segment the drift (Line~\ref{alg:setup:localization:localization:train} in Algorithm~\ref{alg:setup:localization:localization} or Line~\ref{alg:setup:localization:segmentation:train} in Algorithm~\ref{alg:setup:localization:segmentation}) and then analyze it as a proxy to understand the drift as visualized in Figure~\ref{fig:explain:combinations:outlines}. For localization, one can apply any classification model, while for segmentation any segmentation-based multi-regression model can be used. In a second step, one can choose a suitable explanation method for the concrete problem setup. We already presented a range of candidate techniques in Section~\ref{sec:setup:xai}. An inspection and summary on how they can be used in the proposed explanation scheme will be provided in Section~\ref{sec:explain:combinations}. As many explanation approaches are providing explanations for concrete examples, a selection of suitable samples is crucial. We will elaborate on this in the next section. 

A schematic pseudo-code for the explanation routine is given in Algorithm~\ref{alg:explain:method:general} and~\ref{alg:explain:method:local}. The function $\textsc{Preprocess}$ is realized by (drift detection and) timepoint transformation (Line~\ref{alg:setup:localization:localization:preprocess} in Algorithm~\ref{alg:setup:localization:localization} and Line~\ref{alg:setup:localization:segmentation:preprocess} in Algorithm~\ref{alg:setup:localization:segmentation}), the function $\textsc{ExplainModel}$ computes an explanation which is either realized by a global method (Interpretable Model, Global Feature Importance, or Discriminative Dimensionality Reduction) or by Algorithm~\ref{alg:explain:method:local} in case a local explanation (Local Feature Importance, Characteristic Prototypes, Counterfactuals) is applied. In Algorithm~\ref{alg:explain:method:local} characteristic samples (Section~\ref{sec:explain:characteristicsamples}) are computed for each drift segment or region, i.e., all points that belong to the minimal drift locus and are observed before or after the drift. For each prototype obtained this way we compute a local explanation using the respective local explanation method on the model $h$.

\begin{algorithm}[t]
   \caption{Drift Explanation.}
   \label{alg:explain:method:general}
\begin{algorithmic}[1]
   \STATE {\bfseries Input:} {$S = \{(x_1,t_1),...,(x_n,t_n)\}$ dated datapoints}
   \STATE {\bfseries Output:} {$\mathcal{E}$ Explanation}
   \STATE $S' \gets \textsc{Preprocess}(S)$
   \STATE $h \gets \textsc{TrainModel}(S')$
   \STATE $\mathcal{E} \gets \textsc{ExplainModel}(h, \{x_1,\dots,x_n\})$
   \RETURN $\mathcal{E}$
\end{algorithmic}
\end{algorithm}
\begin{algorithm}[t]
   \caption{Local Model Explanation.}
   \label{alg:explain:method:local}
\begin{algorithmic}[1]
   \STATE {\bfseries Input:} {$S = \{x_1,\dots,x_n\}$ datapoints, $h$ trained model}
   \STATE {\bfseries Output:} {$\mathcal{E}$ Explanation}
   \STATE $L \gets \textsc{ExtractSegments}(h, S)$ \COMMENT{Drift Segments or Drift Region}
   \STATE $\mathcal{E} \gets \emptyset$
   \FORALL{Segments $k$ in $L$}
   \STATE $S_k \gets \{x_i \mid L[i] = k \}$
   \STATE $C_k \gets \textsc{ClusterPrototypes}(S_k)$
   \FORALL{Prototpye $c \in C_k$}
   \STATE $\mathcal{E} \gets \mathcal{E} \cup \textsc{ComputeLocalExplanation}(h,c)$
   \ENDFOR
   \ENDFOR
   \RETURN $\mathcal{E}$
\end{algorithmic}
\end{algorithm}

\subsection{Characteristic Samples}
\label{sec:explain:characteristicsamples}
Many recent powerful explanations methods are local, i.e., they require samples that are to be explained in order to be applicable. 
The first step of such approaches is thus to determine the samples which are used for the computation of the explanation. In case there is a human in the loop, they can select the most interesting samples. As we aim for a completely autonomous system that explains the drift at each time step, it is mandatory to automate this step. 
In order to be suited for this task such samples $c_1,\dots,c_n \in \X$ have to fulfill the following requirements:
\begin{enumerate}
    \item they give rise to a partition of the dataspace, i.e., $P : \X \to \{1,\dots,n\}$, $P(c_i) = i$
    \item they represent all associated datapoints, i.e., if $P(x) = P(c)$ then $x$ is represented by $c$
    \item the drifting behavior of the associated samples is homogeneous, i.e., $\P_{T\mid X} \approx \P_{T\mid P(X)}$
\end{enumerate}
We will refer to $c_1,\dots,c_n$ as \emph{characteristic samples}.
As can be seen by the last point the characteristic samples give rise to a drift segmentation. However, as the term ``represent'' is ill-posed and there are several ways how to construct a partition based on prototypes the term characteristic sample is ill-posed, too.

A common way to construct a partition from prototypes is to associate every point in $\X$ with its closest prototype. In this setup, it is reasonable to measure how well a prototype represents a datapoint using the distance between both. Thus we can find characteristic samples by applying prototype-based clustering algorithms like mean shift, Gaussian mixture models, $k$-means, affinity propagation, or spectral clustering. In order to assure the time-homogeneity condition is fulfilled we consider only those samples that belong to one drift segment or drift regions at once, i.e., only those samples that are drifting and are observed before the drift. This approach is presented in Algorithm~\ref{alg:explain:method:local}.

One drawback of this method is that it is hard to control the number of samples per prototype and assure that the clusterings obtained on different subsets are actually compatible. 
One way to solve this issue is to draw ideas from the field of discriminative dimensionality reduction where one considers a metric $d$ on $\X$ that takes model-specific information into account~\cite{DeepView}. As before, by applying this idea to $h(t \mid x)$ to obtain an enriched metric we also capture the drifting behavior. 
To do so we start with a metric on time distributions $d_{\Pr(\T)}$ and then consider the difference in prediction which gives rise to a metric on $\X$: $d_L(x,y) = d_{\Pr(\T)}(h(\cdot\mid x),h(\cdot \mid y))$. This metric captures the drift exactly at the points $x,y$ to capture the geometry of the entire space one uses the length of the shortest path between $x$ and $y$~\cite{DeepView}. 
By applying a prototype-based clustering algorithm using this metric we naturally capture both the geometry of $\X$ which is relevant for representing the data and the drifting behavior. 

A common approach to compute the distance is to start with a $k$-neighbour graph, use $d_L(x,y)+\lambda d_\X(x,y)$ as weights for the edges, and then apply an all pairs shortest path algorithm. 
Unfortunately, this approach is comparably computational expensive. A faster approach is to use MomentTrees as model $h$ and consider the random forest kernel~\cite{ida} of $h$ which captures both the local geometry of $\X$ and the prediction $h(t\mid x)$ at once. 

A more direct approach is to make use of prototype-based models $h$ like LVQ or RBF-networks. The characteristic samples are then given by the prototypes. By definition, those give rise to a time-homogeneous partition. However, in some cases, one has to assure that the prototype actually represents the data. This might require an additional representation loss.

\subsection{Outline: Role of Different Explanation Methods}
\label{sec:explain:combinations}

In the following, we will list inspection and explanation methods and how they can be interpreted and thus used in the described setup. We will point out some potential use cases and demonstrate some of those in Section~\ref{sec:experiments}. We provide an overview in Table~\ref{tab:explain:combinations:overview}. Note that all global explanation methods share the advantage that there is no need to select a characteristic sample.

\begin{table}[!t]
    \centering\small
    \caption{List of explanation schemes.}
    \begin{tabular}{p{2cm}lp{2.5cm}p{2.5cm}lp{0.5cm}}
        \toprule
        Expl. Scheme & Scope & Explains & Output & Runtime & Used \\
        \midrule
        Interpretable Model & global & decision structure of drift & model description & low & -- \\
        Global Feature Importance & global & drifting features & list of features & low & \ref{exp:perturbation}, \ref{exp:bayesnet}, \ref{exp:water} \\
        Discriminative Dimensionality Reduction & global & geometry of drift & scatter plot & medium & \ref{exp:MNIST}\\
        Occurrence Profile & local & local drifting behavior / drift segments & occurrence time diagrams + collection of samples & medium & -- \\
        Local Feature Importance & local & drifting features in subspace & list of samples with marked features & medium & \ref{exp:MNISTPlus} \\
        Counterfactual Explanations & local & drift induced feature alterations & list of contrasting sample pairs & high & \ref{exp:MNIST}, \ref{exp:imagenet} \\
        \bottomrule
    \end{tabular}
    \label{tab:explain:combinations:overview}
\end{table}

\emph{Global Feature Importance}
allows an inspection of the most relevant features for the drift, i.e., those features where the drift manifests. This can be of particular interest in system monitoring when semantic features are used, e.g., if a feature corresponds to a sensor. More broadly speaking it can be used if the system as a whole is of interest, rather than particular parts of the dataspace. A great advantage of this approach is that it does not need to determine characteristic samples, which makes it comparably lightweight. Together with incremental analysis strategies, it can therefore be applied in an online fashion~\cite{fumagalli2022incremental}.
We are going to demonstrate this setup is well suited for the detection and localization of sensor faults in Section~\ref{exp:water}. Furthermore, this approach is applied in the experiments described in Sections~\ref{exp:perturbation}, and \ref{exp:bayesnet}.

\emph{Discrimenative Dimensionality Reduction}
allows for an analysis of the geometry of the drift. This can be relevant if models need to be adjusted by hand or if a global understanding of the drift is necessary. It thus can serve as a first inspection approach. Although we do not need to determine characteristic samples, the procedure can be computationally costly, making a real-time implementation difficult. However, as the provided information can be quite dense, this might not be necessary. 
This scheme is applied in the showcase described in Sections~\ref{exp:MNIST}.

\emph{Occurrence Profile} 
instead of applying a complete explanation scheme, it can also suffice to compute characteristic samples and show them together with the respective occurrence time profile $\P[T \mid X=x]$. This way the process can be monitored in a comparably simple fashion. In particular, this allows to apply more detailed explanations only if requested by the user and is thus comparably computational efficient. 

\emph{Local Feature Importance}
can be of interest if the problem is local in the dataspace. This is for example the case if one is not interested in monitoring the system but rather adjusting learning models. Local feature importances can be obtained by applying LIME or Saliency Maps to the model $h$ in order to provide an explanation for the characteristic samples. This way one obtains an explanation in terms of the most relevant features. In contrast to, for example, permutation feature importances, LIME, and Saliency Maps have been shown to be applicable to high-dimensional, non-semantic data like images. One drawback of this approach is that we only know the most important features. We do not observe the effect of the drift on those. Considering our motivating example (Section~\ref{sec:explain:example}): if a fox is marked as relevant for the drift in a stream of images, is it because the fur color changes or because there are no more foxes in the stream after the drift?
This scheme is applied in the showcase described in Section~\ref{exp:MNISTPlus}.

\emph{Counterfactual Explanations}
are in a sense a perfect match in the case of two timepoints together with drift localization. This is due to the fact that the kind of explanation perfectly matches the type of extracted data: A counterfactual showcases what was changed by the drift in a particular sample. Thus, we can directly observe the effect of the drift. However, it is not necessarily clear how to generalize this to multiple timepoints or drift segmentation as we have to specify the complementary class the counterfactual is supposed to belong to. Natural choices could be any other segment or a counterfactual per segment, etc. %
Furthermore, counterfactuals are usually computationally expensive to obtain and the process is usually not flawless.
This scheme is applied in the showcases described in Sections~\ref{exp:MNIST}, and \ref{exp:imagenet}.

\section{Experiments}
\label{sec:experiments}
We evaluate our methods in several experiments. First, we focus on semantic data (Section~\ref{sec:experiments:global}). Here, global explanation schemes are suitable. However, when considering non-semantic data of high dimensionality, they are not applicable anymore. Thus, we showcase the suitability of local explanations. As an exemplary data domain, we focus on images (Section~\ref{sec:experiments:local}).

\subsection{Global Explanations of Semantic Data\label{sec:experiments:global}}
In order to evaluate the proposed explanation framework with global explanation methods, we present two experiments in which we control the drift in the data. While we investigate relatively simple drift dynamics by inducing drift by feature perturbations in a first experiment (Section~\ref{exp:perturbation}), we consider more complex drift by creating data streams using Bayesian Networks in the second (Section~\ref{exp:bayesnet}). In both experiments, the main goal of the methodology is to identify the drifting features correctly. Finally, in Section~\ref{exp:water} we show how our explanation framework can be applied in critical infrastructure.

\subsubsection{Drift Induced Feature Perturbations} 
\label{exp:perturbation}
\paragraph{Data}
In this experiment, we rely on standard benchmark data and artificially induce drift by perturbing a subset of the features.
We consider the following synthetic datasets:
AGRAWAL~\citep{Agrawal1993DatabaseMA},%
MIXED~\citep{ddm},
RandomRBF~\citep{skmultiflow},
RandomTree~\citep{skmultiflow},
and the following real-world benchmark datasets
``Electricity market prices''~(Elec)~\citep{electricitymarketdata}, 
``Forest Covertype''~(Forest)~\citep{forestcovertypedataset}, and 
``Nebraska Weather'' (Weather)~\citep{weatherdataset}. 
In any case, we consider the joint distribution, i.e., data and label.
To remove uncontrolled effects caused by unknown drift in real-world datasets, we apply a permutation scheme~\citep{ida}. In order to ensure comparability we perform a mean and variance standardization. 
Then, we draw a sub-stream of size 1000 samples from the stream. Abrupt drift is induced at a randomly chosen point between 1/3 and 2/3 of the stream, using one of the following perturbations applied to a varying number (1-5) of randomly selected features: setting to zero, adding a fixed shift of size 1-5, adding standard Gaussian noise, or feature wise permutation of the values.

\paragraph{Setup}
In this experiment, we use the following instantiation of methods for our pipeline. 
We make use of drift segmentation using a Fourier embedding of 5th degree for the time. We consider the following models in a batch and streaming setup, respectively: Decision Tree/Hoeffding Tree (DT), Random Forest/Adaptive Random Forest (RF), Lasso (Las), and Multi Layer Perceptron (MLP; 1-layer, 100 hidden units). To determine the effect of the drift on the feature we make use of the following feature importance measures: permutation feature importance~\cite{breiman2001random} (PFI), incremental permutation feature importance~\cite{fumagalli2022incremental} (IPFI; sum over the entire stream), and feature importance (FI; if available for the model).

\begin{table}[!t]
    \caption{Results of perturbation experiment. Mean over 200 runs and base datasets. }
    \centering
    \tiny
    \begin{minipage}{0.45\textwidth}\centering
\begin{tabular}{l@{\;}l@{\:\:}l@{\:\:}r@{$\pm$}r@{\:\:}r@{$\pm$}r@{\:\:}r@{$\pm$}r@{\:\:}}
\toprule
\multicolumn{2}{c}{Pert.} & & \multicolumn{2}{c}{PFI} & \multicolumn{2}{c}{iPFI} & \multicolumn{2}{c}{FI} \\
\midrule
\multirow{15}{*}{\rotatebox[origin=c]{90}{Constant}}
            & 1 & DT &      0.99 &  0.06 &                  0.91 &  0.20 &     0.78 &  0.22 \\
            &   & Las &      0.50 &  0.32 &                  0.65 &  0.42 & \multicolumn{2}{c}{--} \\
            &   & MLP &      0.95 &  0.17 &                  0.81 &  0.30 & \multicolumn{2}{c}{--} \\
            &   & RF &      0.99 &  0.05 &                  0.85 &  0.25 &     0.79 &  0.22 \\
            & 2 & DT &      0.74 &  0.16 &                  0.86 &  0.19 &     0.51 &  0.22 \\
            &   & Las &      0.50 &  0.25 &                  0.65 &  0.36 & \multicolumn{2}{c}{--} \\
            &   & MLP &      0.90 &  0.15 &                  0.78 &  0.25 & \multicolumn{2}{c}{--} \\
            &   & RF &      0.81 &  0.19 &                  0.85 &  0.20 &     0.51 &  0.26 \\
            & 3 & DT &      0.68 &  0.18 &                  0.82 &  0.19 &     0.40 &  0.23 \\
            &   & Las &      0.50 &  0.22 &                  0.65 &  0.34 & \multicolumn{2}{c}{--} \\
            &   & MLP &      0.85 &  0.16 &                  0.77 &  0.24 & \multicolumn{2}{c}{--} \\
            &   & RF &      0.71 &  0.23 &                  0.84 &  0.19 &     0.40 &  0.28 \\
            & 5 & DT &      0.60 &  0.19 &                  0.81 &  0.18 &     0.27 &  0.21 \\
            &   & Las &      0.50 &  0.18 &                  0.67 &  0.34 & \multicolumn{2}{c}{--} \\
            &   & MLP &      0.77 &  0.18 &                  0.76 &  0.24 & \multicolumn{2}{c}{--} \\
            &   & RF &      0.69 &  0.20 &                  0.82 &  0.19 &     0.30 &  0.27 \\
\midrule
\multirow{15}{*}{\rotatebox[origin=c]{90}{Gaussian Noise}}
            & 1 & DT &      0.92 &  0.21 &                  0.54 &  0.37 &     0.90 &  0.17 \\
            &   & Las &      0.50 &  0.32 &                  0.66 &  0.41 & \multicolumn{2}{c}{--} \\
            &   & MLP &      0.78 &  0.26 &                  0.78 &  0.32 & \multicolumn{2}{c}{--} \\
            &   & RF &      0.99 &  0.07 &                  0.42 &  0.34 &     0.94 &  0.14 \\
            & 2 & DT &      0.78 &  0.20 &                  0.54 &  0.29 &     0.83 &  0.17 \\
            &   & Las &      0.50 &  0.25 &                  0.65 &  0.36 & \multicolumn{2}{c}{--} \\
            &   & MLP &      0.76 &  0.19 &                  0.77 &  0.28 & \multicolumn{2}{c}{--} \\
            &   & RF &      0.87 &  0.17 &                  0.41 &  0.27 &     0.88 &  0.16 \\
            & 3 & DT &      0.72 &  0.20 &                  0.54 &  0.27 &     0.81 &  0.18 \\
            &   & Las &      0.50 &  0.22 &                  0.65 &  0.35 & \multicolumn{2}{c}{--} \\
            &   & MLP &      0.72 &  0.19 &                  0.80 &  0.26 & \multicolumn{2}{c}{--} \\
            &   & RF &      0.79 &  0.19 &                  0.43 &  0.25 &     0.85 &  0.18 \\
            & 5 & DT &      0.64 &  0.18 &                  0.55 &  0.25 &     0.77 &  0.17 \\
            &   & Las &      0.50 &  0.18 &                  0.65 &  0.34 & \multicolumn{2}{c}{--} \\
            &   & MLP &      0.68 &  0.17 &                  0.84 &  0.23 & \multicolumn{2}{c}{--} \\
            &   & RF &      0.69 &  0.19 &                  0.43 &  0.22 &     0.81 &  0.16 \\
\midrule
\multirow{15}{*}{\rotatebox[origin=c]{90}{Value Permutation}}
            & 1 & DT &      0.66 &  0.32 &                  0.58 &  0.34 &     0.62 &  0.33 \\
            &   & Las &      0.51 &  0.31 &                  0.67 &  0.41 & \multicolumn{2}{c}{--} \\
            &   & MLP &      0.59 &  0.33 &                  0.80 &  0.30 & \multicolumn{2}{c}{--} \\
            &   & RF &      0.84 &  0.26 &                  0.43 &  0.31 &     0.65 &  0.34 \\
            & 2 & DT &      0.63 &  0.24 &                  0.58 &  0.29 &     0.60 &  0.25 \\
            &   & Las &      0.51 &  0.25 &                  0.65 &  0.36 & \multicolumn{2}{c}{--} \\
            &   & MLP &      0.55 &  0.23 &                  0.78 &  0.26 & \multicolumn{2}{c}{--} \\
            &   & RF &      0.79 &  0.22 &                  0.44 &  0.26 &     0.63 &  0.26 \\
            & 3 & DT &      0.60 &  0.21 &                  0.58 &  0.27 &     0.57 &  0.23 \\
            &   & Las &      0.49 &  0.21 &                  0.66 &  0.34 & \multicolumn{2}{c}{--} \\
            &   & MLP &      0.53 &  0.22 &                  0.75 &  0.26 & \multicolumn{2}{c}{--} \\
            &   & RF &      0.74 &  0.21 &                  0.45 &  0.25 &     0.62 &  0.23 \\
            & 5 & DT &      0.57 &  0.17 &                  0.61 &  0.25 &     0.56 &  0.18 \\
            &   & Las &      0.50 &  0.18 &                  0.66 &  0.34 & \multicolumn{2}{c}{--} \\
            &   & MLP &      0.51 &  0.18 &                  0.76 &  0.24 & \multicolumn{2}{c}{--} \\
            &   & RF &      0.67 &  0.16 &                  0.49 &  0.23 &     0.59 &  0.18 \\
\bottomrule
\end{tabular}
\end{minipage}
\begin{minipage}{0.45\textwidth}\centering
\begin{tabular}{l@{\;}l@{\:\:}l@{\:\:}r@{$\pm$}r@{\:\:}r@{$\pm$}r@{\:\:}r@{$\pm$}r@{\:\:}}
\toprule
\multicolumn{2}{c}{Pert.} & & \multicolumn{2}{c}{PFI} & \multicolumn{2}{c}{iPFI} & \multicolumn{2}{c}{FI} \\
\midrule
\multirow{15}{*}{\rotatebox[origin=c]{90}{Shift (+1)}}
            & 1 & DT &      0.99 &  0.05 &                  0.98 &  0.10 &     0.83 &  0.21 \\
            &   & Las &      0.49 &  0.31 &                  0.64 &  0.42 & \multicolumn{2}{c}{--} \\
            &   & MLP &      0.99 &  0.05 &                  0.86 &  0.25 & \multicolumn{2}{c}{--} \\
            &   & RF &      1.00 &  0.00 &                  0.87 &  0.26 &     0.87 &  0.20 \\
            & 2 & DT &      0.89 &  0.15 &                  0.98 &  0.09 &     0.75 &  0.21 \\
            &   & Las &      0.50 &  0.24 &                  0.64 &  0.36 & \multicolumn{2}{c}{--} \\
            &   & MLP &      0.87 &  0.15 &                  0.82 &  0.22 & \multicolumn{2}{c}{--} \\
            &   & RF &      0.89 &  0.17 &                  0.84 &  0.23 &     0.79 &  0.21 \\
            & 3 & DT &      0.81 &  0.17 &                  0.97 &  0.09 &     0.70 &  0.21 \\
            &   & Las &      0.51 &  0.22 &                  0.65 &  0.35 & \multicolumn{2}{c}{--} \\
            &   & MLP &      0.80 &  0.18 &                  0.79 &  0.24 & \multicolumn{2}{c}{--} \\
            &   & RF &      0.82 &  0.19 &                  0.82 &  0.21 &     0.74 &  0.22 \\
            & 5 & DT &      0.71 &  0.17 &                  0.94 &  0.11 &     0.62 &  0.19 \\
            &   & Las &      0.49 &  0.19 &                  0.67 &  0.34 & \multicolumn{2}{c}{--} \\
            &   & MLP &      0.74 &  0.18 &                  0.80 &  0.22 & \multicolumn{2}{c}{--} \\
            &   & RF &      0.68 &  0.19 &                  0.76 &  0.18 &     0.69 &  0.20 \\
\midrule
\multirow{15}{*}{\rotatebox[origin=c]{90}{Shift (+2)}}
            & 1 & DT &      1.00 &  0.00 &                  0.99 &  0.09 &     0.86 &  0.20 \\
            &   & Las &      0.51 &  0.33 &                  0.64 &  0.42 & \multicolumn{2}{c}{--} \\
            &   & MLP &      1.00 &  0.01 &                  0.91 &  0.21 & \multicolumn{2}{c}{--} \\
            &   & RF &      1.00 &  0.00 &                  0.94 &  0.21 &     0.87 &  0.19 \\
            & 2 & DT &      0.95 &  0.12 &                  0.99 &  0.07 &     0.77 &  0.20 \\
            &   & Las &      0.50 &  0.24 &                  0.64 &  0.36 & \multicolumn{2}{c}{--} \\
            &   & MLP &      0.94 &  0.10 &                  0.84 &  0.22 & \multicolumn{2}{c}{--} \\
            &   & RF &      0.95 &  0.13 &                  0.93 &  0.18 &     0.79 &  0.21 \\
            & 3 & DT &      0.88 &  0.15 &                  0.98 &  0.07 &     0.71 &  0.20 \\
            &   & Las &      0.50 &  0.22 &                  0.63 &  0.35 & \multicolumn{2}{c}{--} \\
            &   & MLP &      0.91 &  0.11 &                  0.81 &  0.23 & \multicolumn{2}{c}{--} \\
            &   & RF &      0.86 &  0.18 &                  0.85 &  0.19 &     0.76 &  0.19 \\
            & 5 & DT &      0.74 &  0.16 &                  0.96 &  0.08 &     0.63 &  0.18 \\
            &   & Las &      0.50 &  0.19 &                  0.66 &  0.34 & \multicolumn{2}{c}{--} \\
            &   & MLP &      0.87 &  0.14 &                  0.80 &  0.23 & \multicolumn{2}{c}{--} \\
            &   & RF &      0.67 &  0.19 &                  0.72 &  0.20 &     0.71 &  0.19 \\
\midrule
\multirow{15}{*}{\rotatebox[origin=c]{90}{Shift (+5)}}
            & 1 & DT &      1.00 &  0.00 &                  0.99 &  0.07 &     0.90 &  0.17 \\
            &   & Las &      1.00 &  0.00 &                  0.65 &  0.42 & \multicolumn{2}{c}{--} \\
            &   & MLP &      1.00 &  0.00 &                  0.95 &  0.16 & \multicolumn{2}{c}{--} \\
            &   & RF &      1.00 &  0.00 &                  0.94 &  0.21 &     0.91 &  0.16 \\
            & 2 & DT &      0.75 &  0.17 &                  0.99 &  0.07 &     0.70 &  0.20 \\
            &   & Las &      0.96 &  0.11 &                  0.64 &  0.37 & \multicolumn{2}{c}{--} \\
            &   & MLP &      1.00 &  0.01 &                  0.88 &  0.19 & \multicolumn{2}{c}{--} \\
            &   & RF &      0.80 &  0.20 &                  0.94 &  0.16 &     0.79 &  0.19 \\
            & 3 & DT &      0.67 &  0.18 &                  0.98 &  0.09 &     0.63 &  0.19 \\
            &   & Las &      0.90 &  0.13 &                  0.66 &  0.34 & \multicolumn{2}{c}{--} \\
            &   & MLP &      1.00 &  0.01 &                  0.81 &  0.25 & \multicolumn{2}{c}{--} \\
            &   & RF &      0.63 &  0.22 &                  0.89 &  0.19 &     0.73 &  0.20 \\
            & 5 & DT &      0.60 &  0.16 &                  0.95 &  0.12 &     0.59 &  0.17 \\
            &   & Las &      0.82 &  0.14 &                  0.64 &  0.35 & \multicolumn{2}{c}{--} \\
            &   & MLP &      1.00 &  0.01 &                  0.80 &  0.23 & \multicolumn{2}{c}{--} \\
            &   & RF &      0.50 &  0.18 &                  0.80 &  0.20 &     0.69 &  0.18 \\
\bottomrule
\end{tabular}
\end{minipage}
    \label{tab:exp:perturbation:results:short}
\end{table}

We evaluate the results using an AUC-ROC score, i.e., rank the features according to the respective importance score and then check how well the ranking aligns with whether a feature is (non-) drifting. We repeated the process 200 times. 

\paragraph{Results}
A summary of the results is shown in Table~\ref{tab:exp:perturbation:results:short}.%

As can be seen, except for the value permutation, the number of affected features has the strongest effect on the performance (the more the harder), followed by the used inspection method and model. The effect of the used dataset is nearly negligible, the effect of the used perturbation depends: it is very similar for setting to zero (Constant) and adding Gaussian Noise, and the Shifts with different intensities which seem to become easier for larger shifts, value permutation appears to be the hardest problem. 

Regarding model and inspection method we observe that DT and RF usually perform rather comparably, although RF works better with PFI and FI, whereas DT appears to be more compatible with iPFI. However, this can be caused by both the inspection method and the stream learner. For Las, we observe that for PFI the results are equivalent to random chance, which are consistently outperformed by iPFI. This could be explained by the fact that the problem cannot be learned by a linear model, which is still capable of learning a smaller time window. For DT, RF, and MLP this consideration is inconclusive regarding the mean, but iPFI usually shows a larger variance. 

To conclude, the number of affected features has the strongest effect on the result, all other parameters are either negligible or inconclusive. In particular, the incremental approach (iPFI) is not outperformed by the batch methods.

\subsubsection{Drifting Bayesian Networks}
\label{exp:bayesnet}
\begin{figure}[!t]
    \centering
    \begin{minipage}{0.40\textwidth}
        \centering
        \bayesTicks
        \subcaption{Illustration of random Bayes-Network. Directly time-affected nodes are connected to a $T$-node ($I_1$, $I_2$). Drifting nodes are marked with a thick border line ($I_1$-$I_4$, $F_1$-$F_7$), non-drifting nodes are marked with a thin border line ($N_1$-$N_3$). Nodes with dashed border line are only present in the ``complete'' setup ($F_1$-$F_7$).}
        \label{fig:exp:bayesnet:net}
    \end{minipage}
    \hfill
    \begin{minipage}{0.55\textwidth} \centering
        \includegraphics[width=0.45\textwidth]{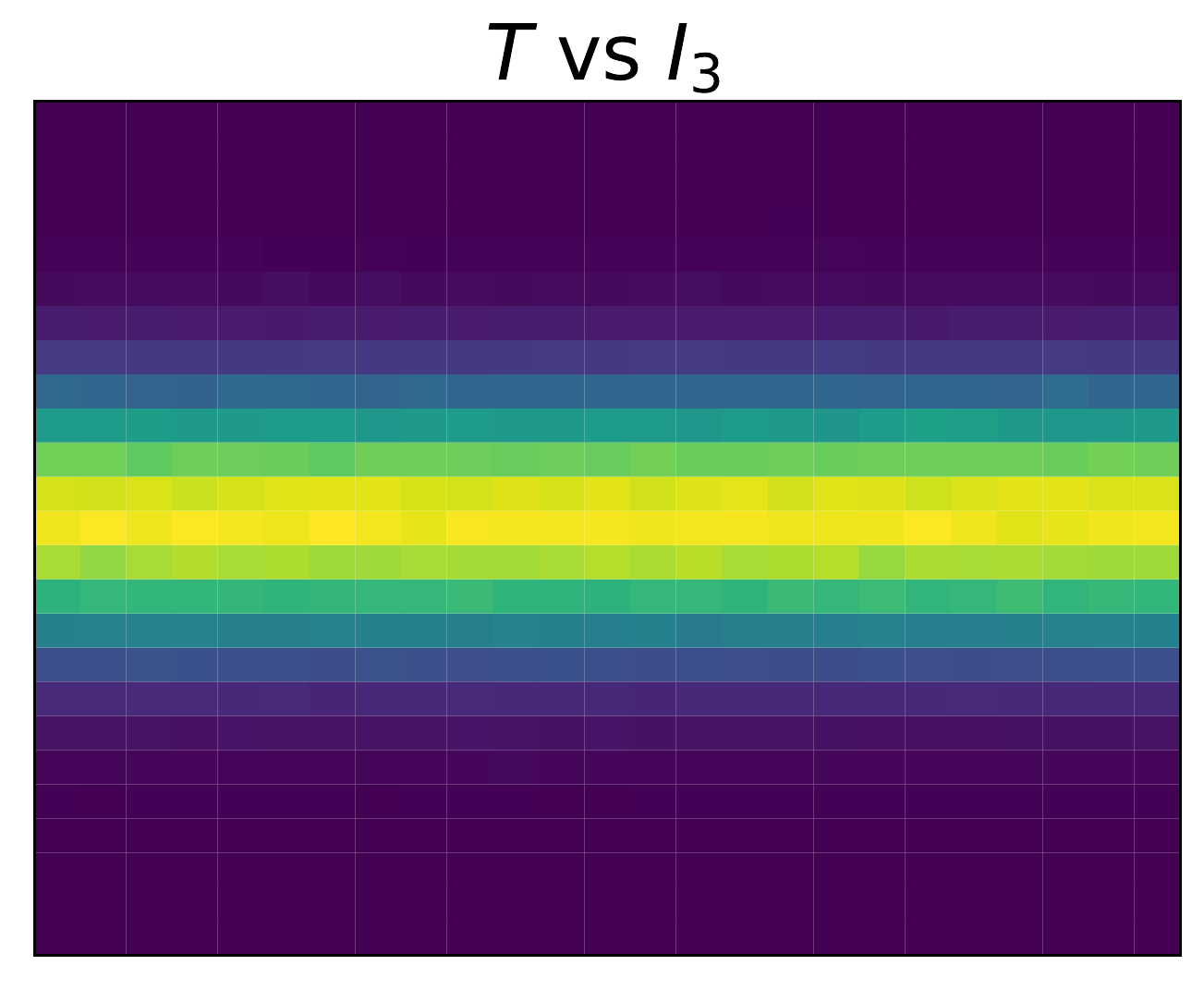}
        \includegraphics[width=0.45\textwidth]{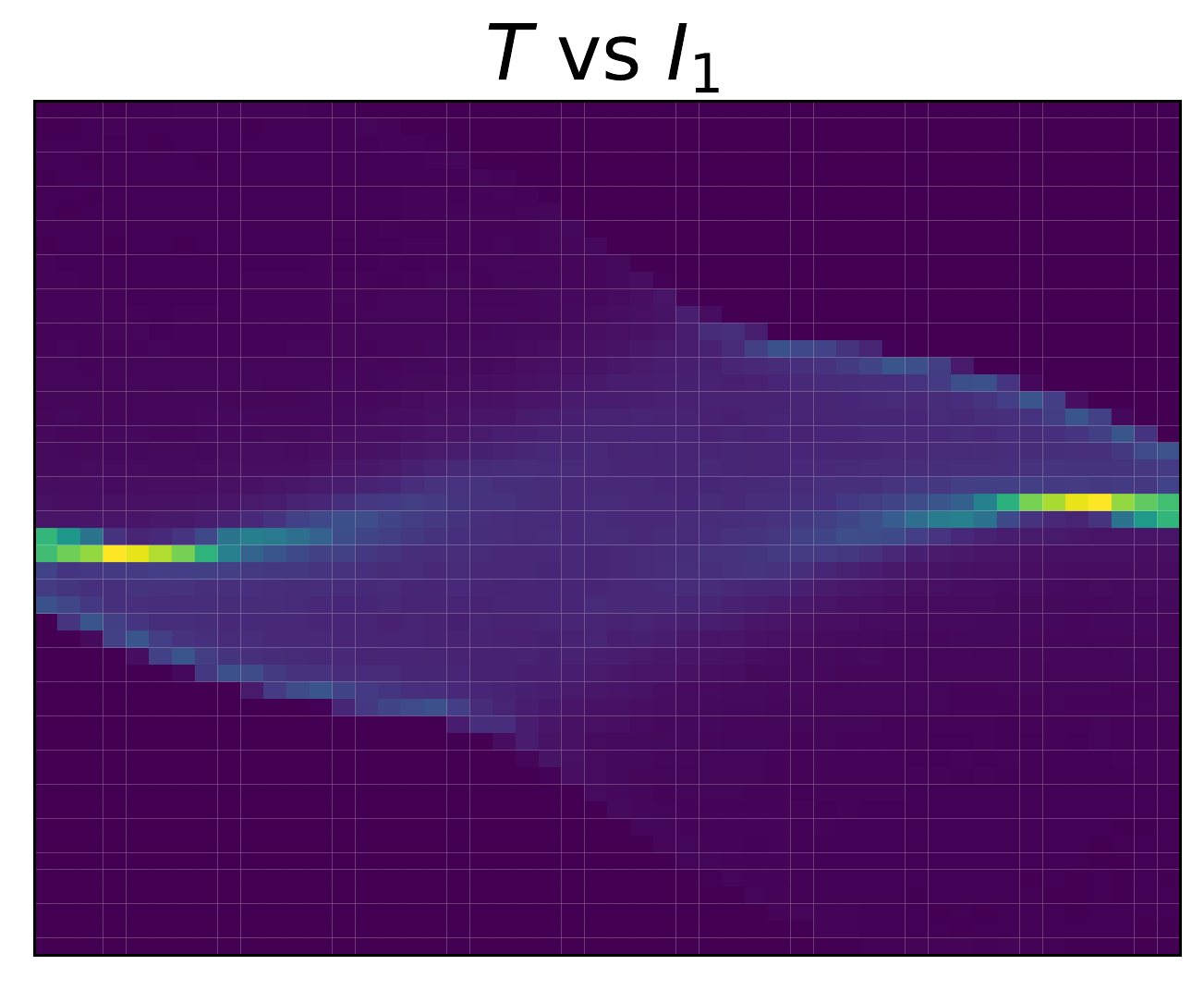}
        \includegraphics[width=0.45\textwidth]{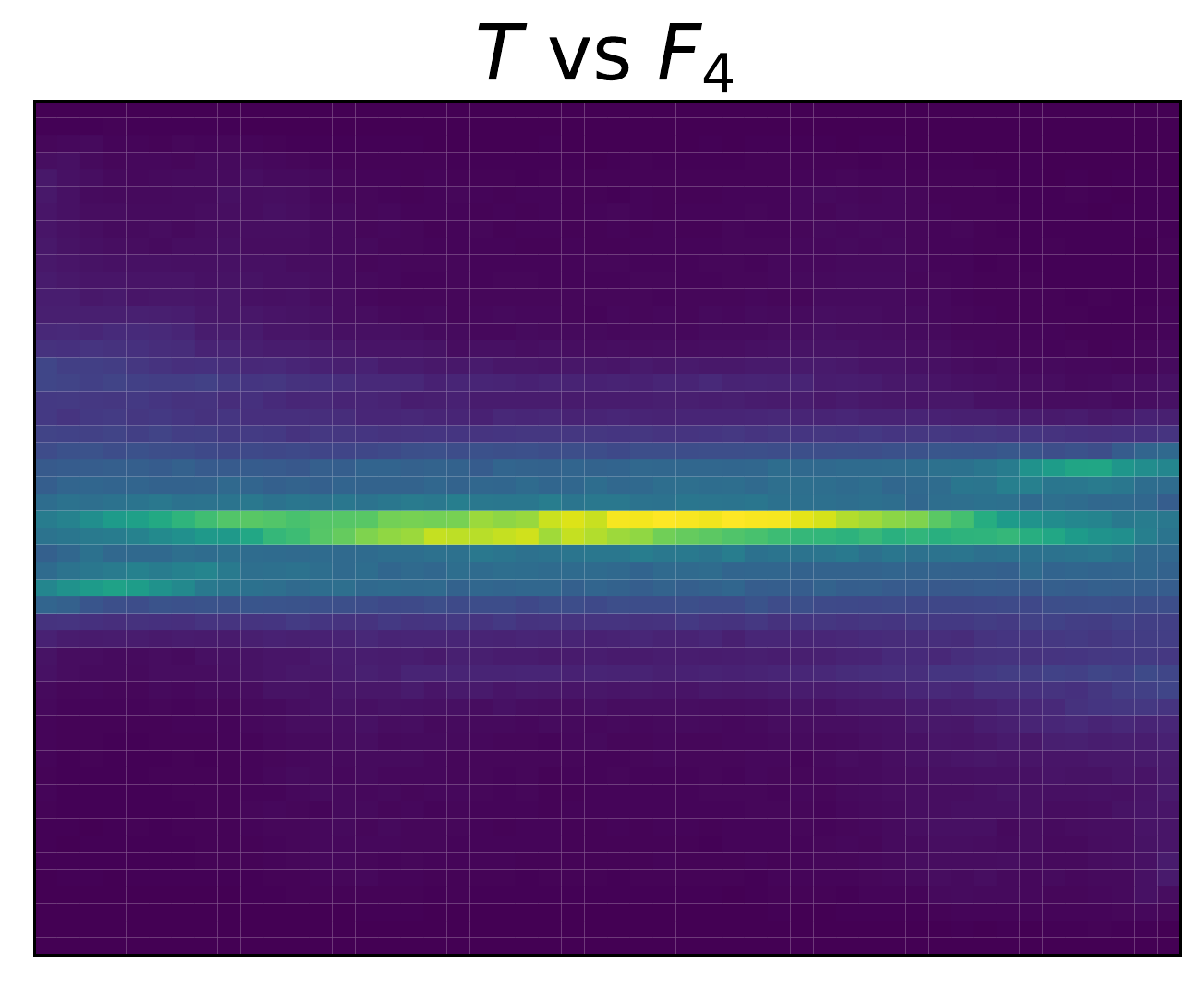}
        \includegraphics[width=0.45\textwidth]{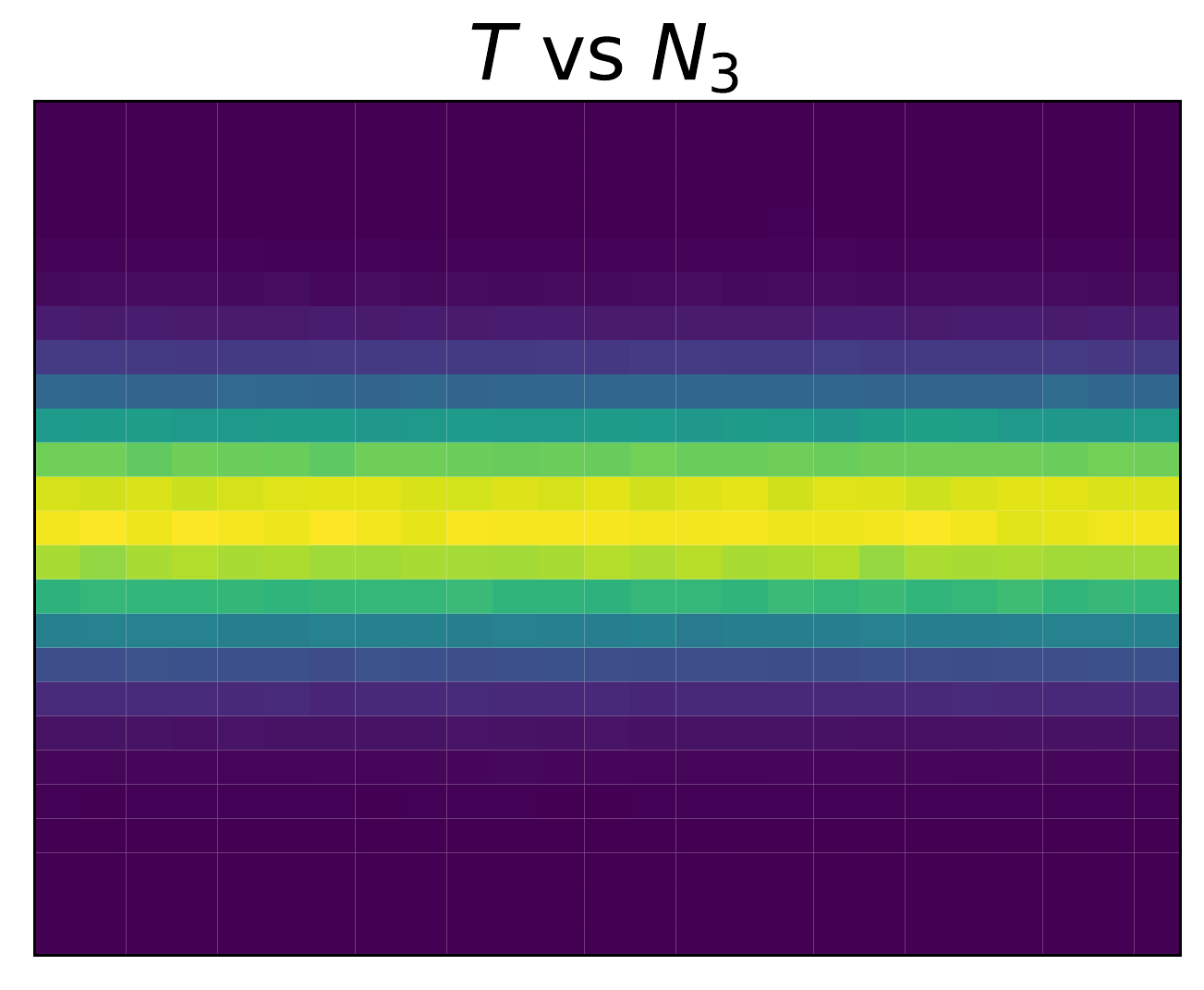} \\ 
        \includegraphics[width=0.45\textwidth]{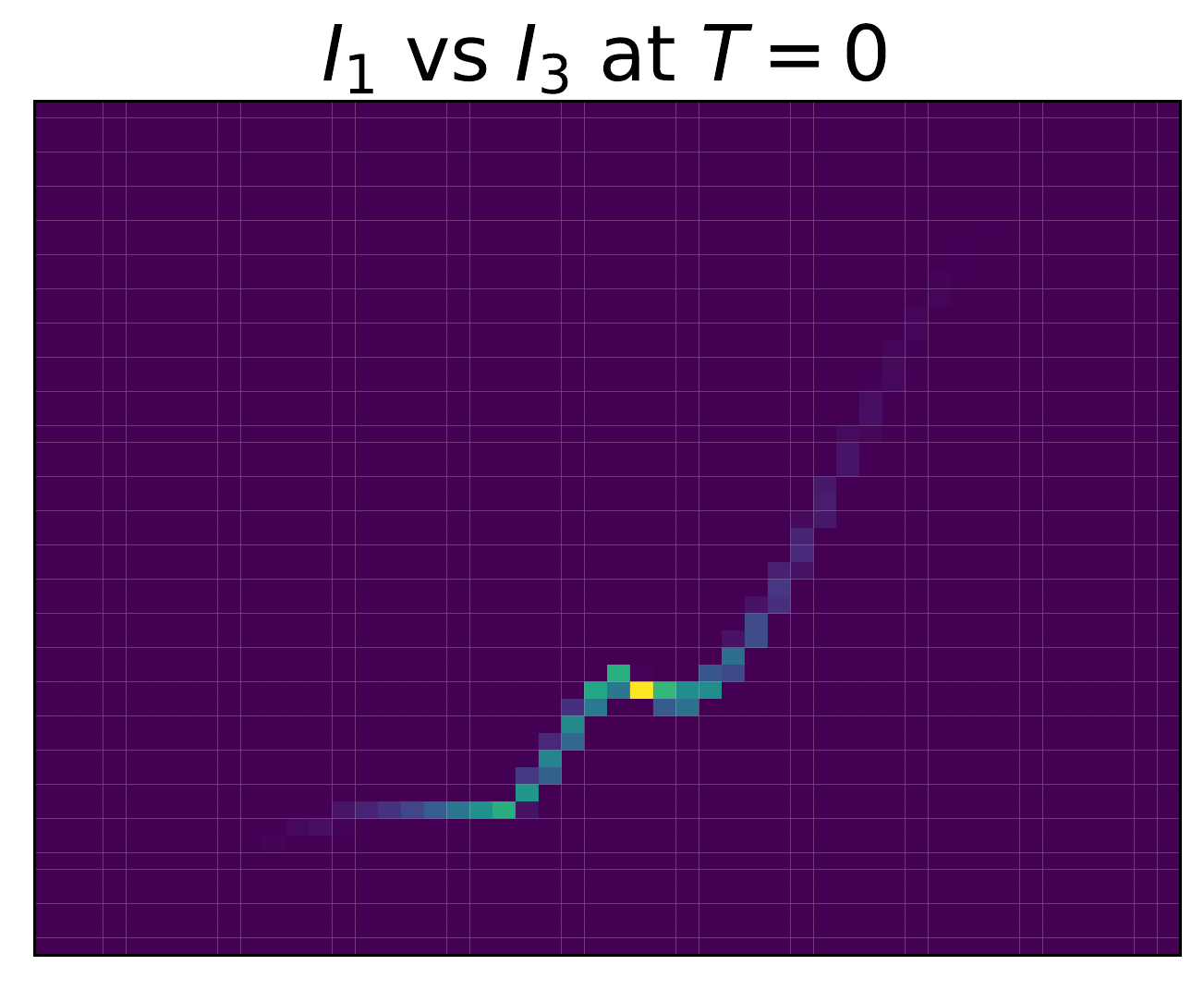}
        \includegraphics[width=0.45\textwidth]{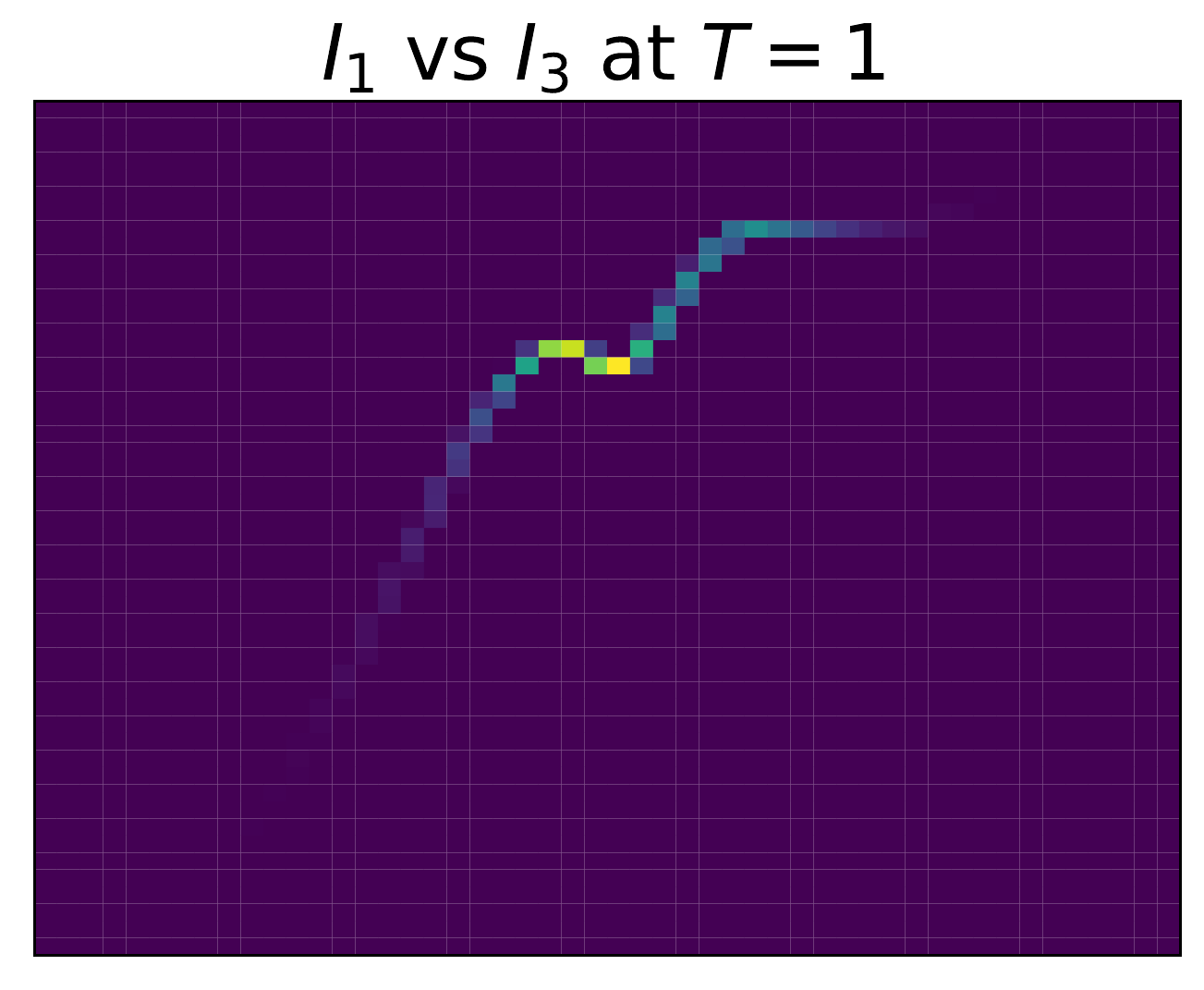}
        \subcaption{Illustration of distribution generated by the network. Upper four pictures show time ($T$) on $x$-axis and feature value ($I_3,I_1,F_1,N_3$) on $y$-axis. Lower two show feature value of $I_3$ on $x$-axis and $I_1$ on $y$-axis for different timepoint (left: $T=0$, right: $T=1$). As can be seen time has a strong effect on the value of $I_1$ and the correlation of $I_1$ and $I_3$, a small effect on $F_4$, and no effect on $I_3$ (on its own) and $N_3$.
        }
        \label{fig:exp:bayesnet:distribution}
    \end{minipage}
    \caption{Illustration of Bayes-Network data.}
    \label{fig:exp:bayesnet}
\end{figure}

While the last experiment demonstrated that the proposed explanation framework works for simple abrupt drift dynamics, we aim to present its suitability to more complex drift dynamics in this experiment.
\paragraph{Data}
To generate data streams with more complex drift behavior, we consider randomly constructed Bayesian networks, like the one visualized in Figure~\ref{fig:exp:bayesnet}. Each node takes on a normal distribution, where mean and variance are computed using randomly initialized neural networks. Drift is introduced by making the distribution of some of the nodes time-dependent, i.e., $X_v \mid X_{\text{pa}(v)} \sim p_t(X_v \mid  X_{\text{pa}(v)})$. This is realized by making time one of the input features of the network. We illustrated the resulting distribution for some features over time in Figure~\ref{fig:exp:bayesnet:distribution}. 

Clearly, all nodes that are directly affected by the time are drifting as well as their children, but also the parent nodes of those as the correlation between the features is affected by the drift. By similar arguments, we inductively obtain all nodes that are in the same connected component as one of the nodes that directly depends on time as drifting. However, it is reasonable to assume that those nodes that are further away from the ones that are directly affected by the drift are ``less drifting''. To evaluate this effect we run this experiment in two modes. While we generate data based on the entire network in the ``complete'' setting, we only consider the sub-networks that only consist of nodes that are directly affected by the drift, their parents, and the non-drifting nodes as the ``shallow'' setup. In Figure~\ref{fig:exp:bayesnet:net} the nodes not contained in the shallow setup are marked by a dashed line. 

\begin{table}[!t]
    \caption{Analysis on random Bayesian networks. Mean over 200 runs. Desc. is number of features directly/implicitly/not affected by drift.}
    \centering\scriptsize
    \begin{tabular}{l@{\:\:}lr@{$\pm$}r@{\:\:}r@{$\pm$}r@{\:\:}r@{$\pm$}r@{\qquad}r@{$\pm$}r@{\:\:}r@{$\pm$}r@{\:\:}r@{$\pm$}r}
\toprule
  &  & \multicolumn{6}{c}{Complete} & \multicolumn{6}{c}{Shallow} \\
Desc. & Model & \multicolumn{2}{l}{PFI} & \multicolumn{2}{l}{iPFI} & \multicolumn{2}{l}{FI} & \multicolumn{2}{l}{PFI} & \multicolumn{2}{l}{iPFI} & \multicolumn{2}{l}{FI} \\
\midrule
\multirow{4}{*}{\rotatebox[origin=c]{90}{5/15/5}}
  & DT &        0.61 &  0.12 &                    0.20 &  0.10 &       0.59 &  0.12 &        0.87 &  0.10 &                    0.74 &  0.17 &       0.84 &  0.09 \\
  & Las &        0.51 &  0.15 &                    0.94 &  0.16 & \multicolumn{2}{c}{--}  &        0.49 &  0.19 &                    0.99 &  0.05 & \multicolumn{2}{c}{--}  \\
  & MLP &        0.80 &  0.07 &                    0.11 &  0.13 & \multicolumn{2}{c}{--}  &        1.00 &  0.00 &                    0.55 &  0.19 & \multicolumn{2}{c}{--}  \\
  & RF &        0.64 &  0.12 &                    0.66 &  0.12 &       0.61 &  0.11 &        0.87 &  0.10 &                    0.99 &  0.03 &       0.88 &  0.07 \\
\midrule
\multirow{4}{*}{\rotatebox[origin=c]{90}{7/13/5}}
  & DT &        0.65 &  0.11 &                    0.08 &  0.05 &       0.55 &  0.11 &        0.86 &  0.09 &                    0.24 &  0.12 &       0.84 &  0.10 \\
  & Las &        0.51 &  0.15 &                    0.59 &  0.37 & \multicolumn{2}{c}{--}  &        0.49 &  0.19 &                    0.81 &  0.24 & \multicolumn{2}{c}{--}  \\
  & MLP &        0.84 &  0.07 &                    0.28 &  0.12 & \multicolumn{2}{c}{--}  &        0.72 &  0.05 &                    0.62 &  0.17 & \multicolumn{2}{c}{--}  \\
  & RF &        0.66 &  0.11 &                    0.62 &  0.14 &       0.53 &  0.07 &        0.90 &  0.08 &                    0.87 &  0.09 &       0.90 &  0.05 \\
\midrule
\multirow{4}{*}{\rotatebox[origin=c]{90}{6/11/8}}
  & DT &        0.58 &  0.11 &                    0.59 &  0.15 &       0.55 &  0.09 &        0.70 &  0.14 &                    0.87 &  0.11 &       0.79 &  0.09 \\
  & Las &        0.50 &  0.14 &                    0.99 &  0.02 & \multicolumn{2}{c}{--}  &        0.50 &  0.15 &                    1.00 &  0.01 & \multicolumn{2}{c}{--}  \\
  & MLP &        0.69 &  0.09 &                    0.19 &  0.06 & \multicolumn{2}{c}{--}  &        0.63 &  0.06 &                    0.51 &  0.11 & \multicolumn{2}{c}{--}  \\
  & RF &        0.59 &  0.13 &                    0.49 &  0.13 &       0.56 &  0.06 &        0.84 &  0.11 &                    0.59 &  0.11 &       0.86 &  0.04 \\
\midrule
\multirow{4}{*}{\rotatebox[origin=c]{90}{6/14/5}}
  & DT &        0.66 &  0.12 &                    0.55 &  0.17 &       0.39 &  0.10 &        0.85 &  0.10 &                    0.93 &  0.08 &       0.61 &  0.10 \\
  & Las &        0.50 &  0.15 &                    0.32 &  0.33 & \multicolumn{2}{c}{--}  &        0.49 &  0.17 &                    0.39 &  0.31 & \multicolumn{2}{c}{--}  \\
  & MLP &        0.83 &  0.08 &                    0.15 &  0.12 & \multicolumn{2}{c}{--}  &        0.71 &  0.08 &                    0.46 &  0.21 & \multicolumn{2}{c}{--}  \\
  & RF &        0.72 &  0.12 &                    0.45 &  0.16 &       0.33 &  0.06 &        0.91 &  0.08 &                    0.86 &  0.02 &       0.61 &  0.07 \\
\midrule
\multirow{4}{*}{\rotatebox[origin=c]{90}{6/8/11}}
  & DT &        0.60 &  0.10 &                    0.70 &  0.14 &       0.64 &  0.10 &        0.74 &  0.10 &                    0.80 &  0.11 &       0.81 &  0.10 \\
  & Las &        0.50 &  0.12 &                    0.85 &  0.24 & \multicolumn{2}{c}{--}  &        0.51 &  0.16 &                    0.88 &  0.20 & \multicolumn{2}{c}{--}  \\
  & MLP &        0.57 &  0.06 &                    0.32 &  0.09 & \multicolumn{2}{c}{--}  &        0.56 &  0.05 &                    0.59 &  0.12 & \multicolumn{2}{c}{--}  \\
  & RF &        0.62 &  0.11 &                    0.50 &  0.12 &       0.68 &  0.08 &        0.78 &  0.11 &                    0.57 &  0.11 &       0.87 &  0.07 \\
\midrule
\multirow{4}{*}{\rotatebox[origin=c]{90}{6/7/12}}
  & DT &        0.64 &  0.10 &                    0.16 &  0.12 &       0.62 &  0.09 &        0.74 &  0.12 &                    0.80 &  0.13 &       0.75 &  0.10 \\
  & Las &        0.52 &  0.11 &                    0.67 &  0.34 & \multicolumn{2}{c}{--}  &        0.49 &  0.15 &                    0.79 &  0.26 & \multicolumn{2}{c}{--}  \\
  & MLP &        0.72 &  0.06 &                    0.36 &  0.16 & \multicolumn{2}{c}{--}  &        0.76 &  0.08 &                    0.53 &  0.13 & \multicolumn{2}{c}{--}  \\
  & RF &        0.64 &  0.11 &                    0.62 &  0.11 &       0.70 &  0.07 &        0.79 &  0.11 &                    0.74 &  0.11 &       0.80 &  0.08 \\
\bottomrule
\end{tabular}
    \label{tab:exp:bayesnet}
\end{table}

\paragraph{Setup}
In this experiment, we apply the same methodology as in the last experiment (Section~\ref{exp:perturbation})

\paragraph{Results}
The results are shown in Table~\ref{tab:exp:bayesnet}.
As can be seen, the shallow network is easier to handle than the complete one. If we consider only the nodes that are directly affected by the drift and their parents as drifting, this difference vanishes. Furthermore, MLP+PFI and Las+iPFI perform best on the complete setup, RF+PFI, DT+iPFI, and Las+iPFI on the shallow setup. MLP and RF are more compatible with PFI than iPFI. For RF and DT, in the complete PFI usually outperforms FI, on the shallow setup this is inconclusive. RF and DT with PFI and FI are usually comparable.
To conclude, Las+iPFI performs surprisingly well on many datasets. For the more complex models, MLP and RF are less compatible with iPFI than PFI.

\subsubsection{Detection and Localization of Sensor Faults in Water Distribution Networks}
\label{exp:water}

\begin{figure}[!t]
    \centering
    \begin{minipage}{0.45\textwidth}
    \includegraphics[width=0.9\textwidth]{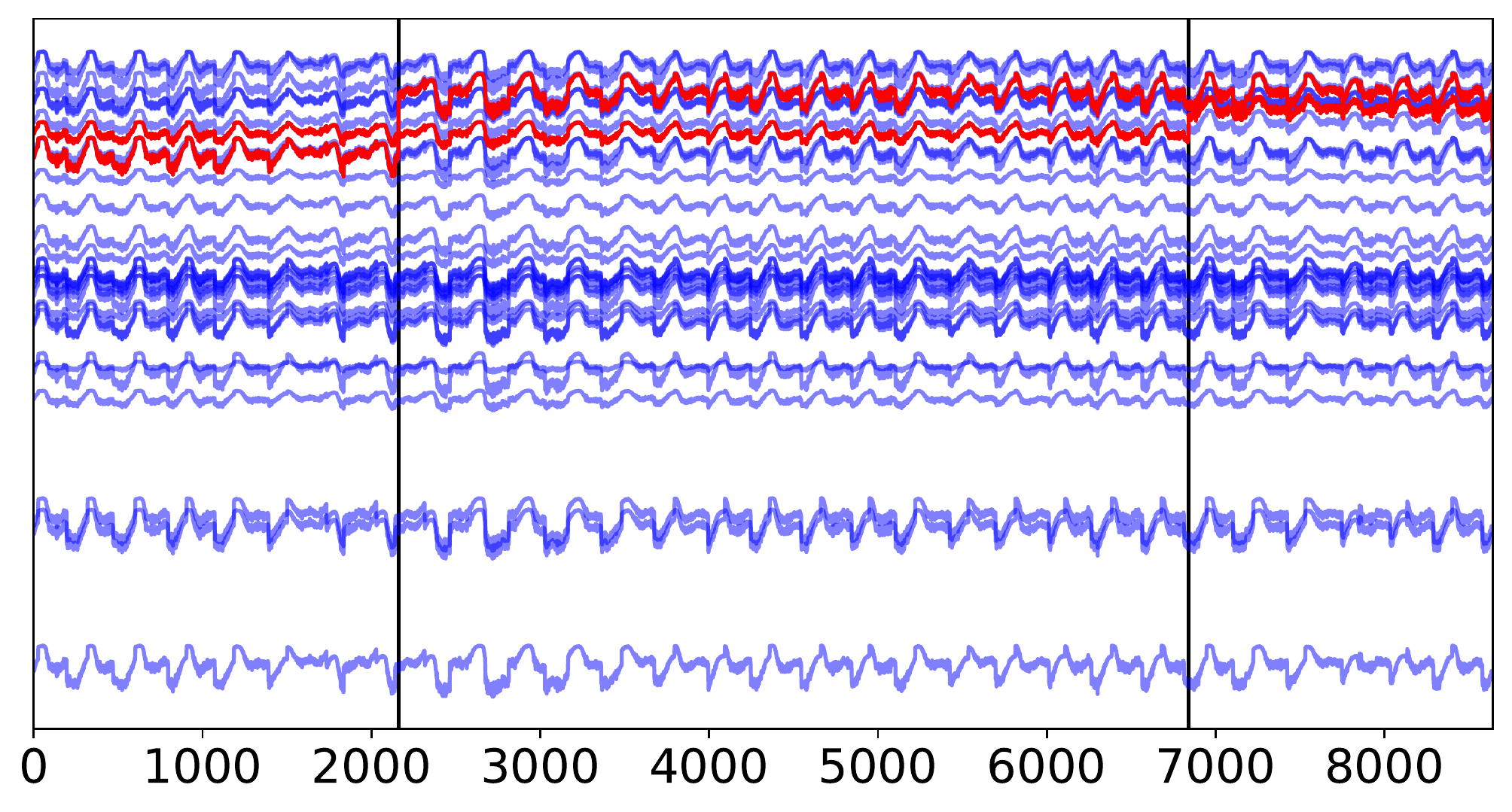}
    \subcaption{Raw data}
    \label{fig:exp:water:data}
    \end{minipage}
    \begin{minipage}{0.45\textwidth}
    \includegraphics[width=0.9\textwidth]{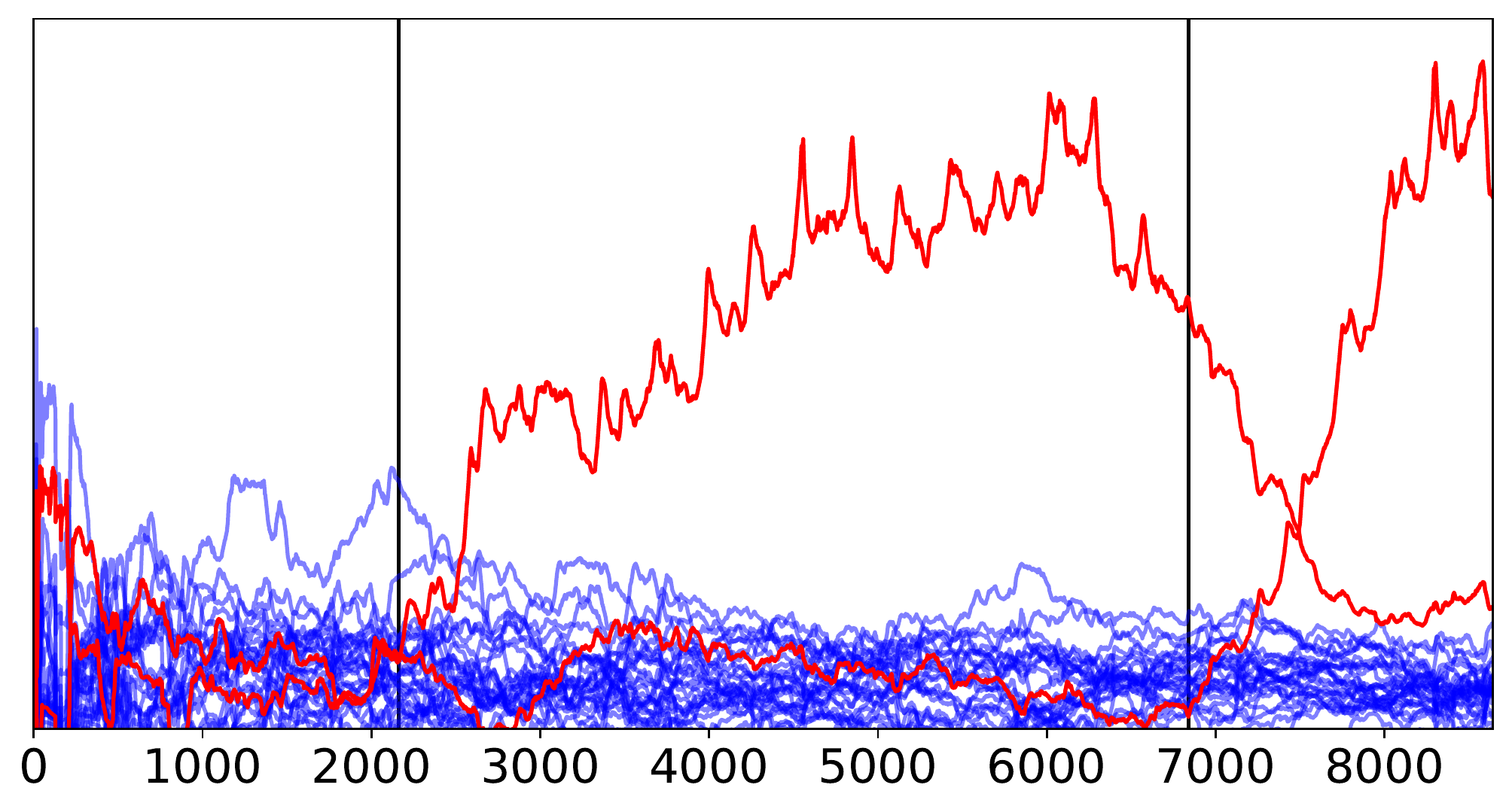}
    \subcaption{Incremental permutation feature importance}
    \label{fig:exp:water:importances}
    \end{minipage}
    \caption{Results of (incremental) feature analysis of water data. Plots show: timepoints of fault (black lines), unaffected features (blue lines), affected features (red lines)}
    \label{fig:exp:water}
\end{figure}
Finally, we present a real-world use case for global explanations of concept drift.
Critical infrastructures like water distribution networks are usually monitored by several sensors that continuously take  measurements of the system. Changes in the reported values indicate a change in the system state that might require manual intervention to assure the integrity of the system or prevent malfunctions. However, as the monitoring system itself can be affected by malfunctions, such as sensor fault, it is important to detect and identify those, i.e., determine the timepoint when the fault happens and identify the broken sensor(s)~\citep{DBLP:conf/icann/VaquetABH22}. This task is non-trivial as the sensor readings are also affected by changing consumer demands which are in turn affected by external factors like the day-night-cycle, workday-and-weekend as well as public holidays, large sports events, or the current weather situation, which also result in complex changes in the sensor readings~\citep{vonk_estimating_2019}. 

\paragraph{Data}
We generated a dataset of pressure values in the L-Town network using realistic demands \citep{battledim} for a water distribution network using a commonly used simulation tool from the literature~\citep{wntr}. We add two sensor faults at different timepoints (see Figure~\ref{fig:exp:water:data}).

\paragraph{Setup}
As each feature corresponds to a single sensor we address the task of sensor fault localization by means of feature importance. We use an adaptive random forest as a model for drift segmentation using a Fourier transform of degree 5 and a period of 500 samples. We analyze the model using incremental permutation feature importances~\cite{fumagalli2022incremental}. 

\paragraph{Results}
The results are presented in Figure~\ref{fig:exp:water:importances}. As can be seen, the method correctly identifies the faulty sensors, albeit with some delay. This is consistent with the findings in the first experiment (Section~\ref{exp:perturbation}) and we were able to obtain similar results on data generated using different sensors and faults. %

\subsection{Local Explanations on High-Dimensional Non-Semantic Data}
\label{sec:experiments:local}
So far we have focused on global explanations and semantic data. Such explanations are comparably simple as it suffices to point out relevant features, either by directly marking them or by showing suitable groupings to point out relevant correlations. However, many real-world data sources do not provide a direct, feature-wise interpretation -- especially when they are lacking clear semantics and are high-dimensional. In this section, we will consider image data as an exemplary data domain and explore potential ways to extract drift-related information by local explanation schemes. First, we will focus on rather simple MNIST based scenarios (Section~\ref{exp:MNISTPlus}, \ref{exp:MNIST}). Afterward, we will show an example application of the proposed explanation methodology on a more complex data stream (Section\ref{exp:imagenet}).

\subsubsection{Local Feature-Based Explanations for Non-Sematic Data using MNIST}
\label{exp:MNISTPlus}

\begin{figure}[!t]
    \centering
    \begin{minipage}{0.325\textwidth}
    \centering
    \includegraphics[width=0.95\textwidth]{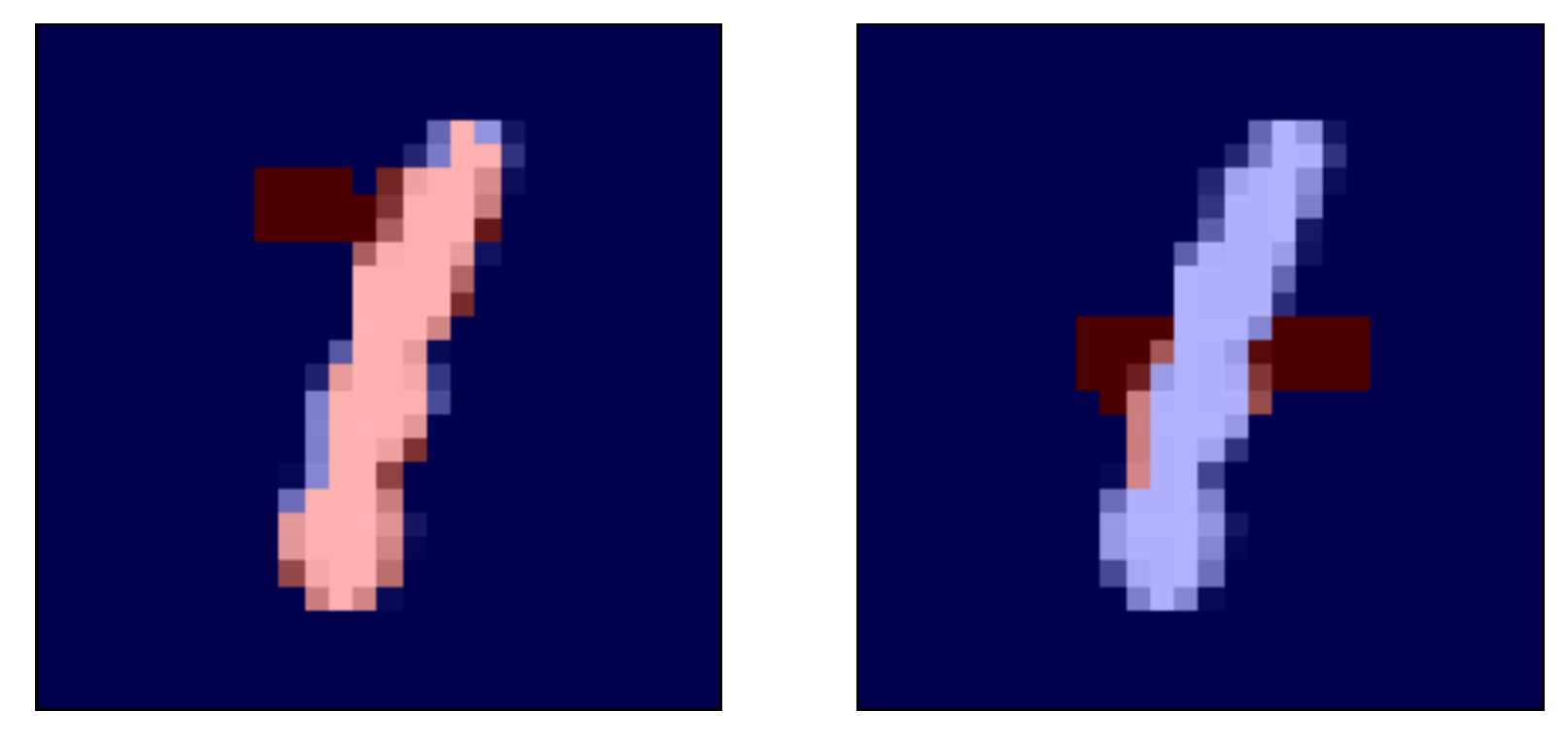}
    \subcaption{Sample from before drift. \\ \;}
    \label{fig:exp:MNISTPlus:1}
    \end{minipage}
    \begin{minipage}{0.325\textwidth}
    \centering
    \includegraphics[width=0.95\textwidth]{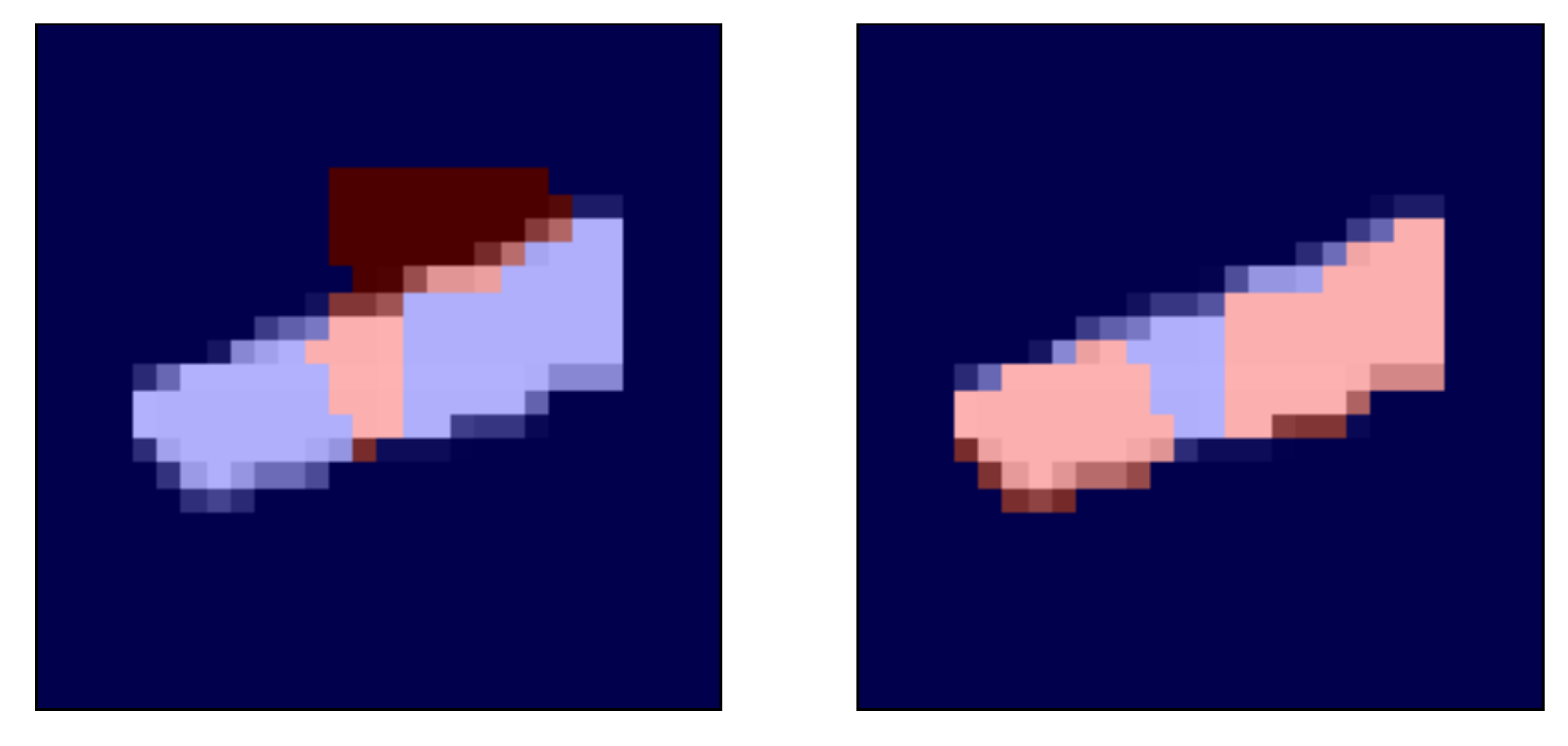}
    \subcaption{Sample from after drift.\\ \;}
    \label{fig:exp:MNISTPlus:-}
    \end{minipage}
    \begin{minipage}{0.325\textwidth}
    \centering
    \includegraphics[width=0.95\textwidth]{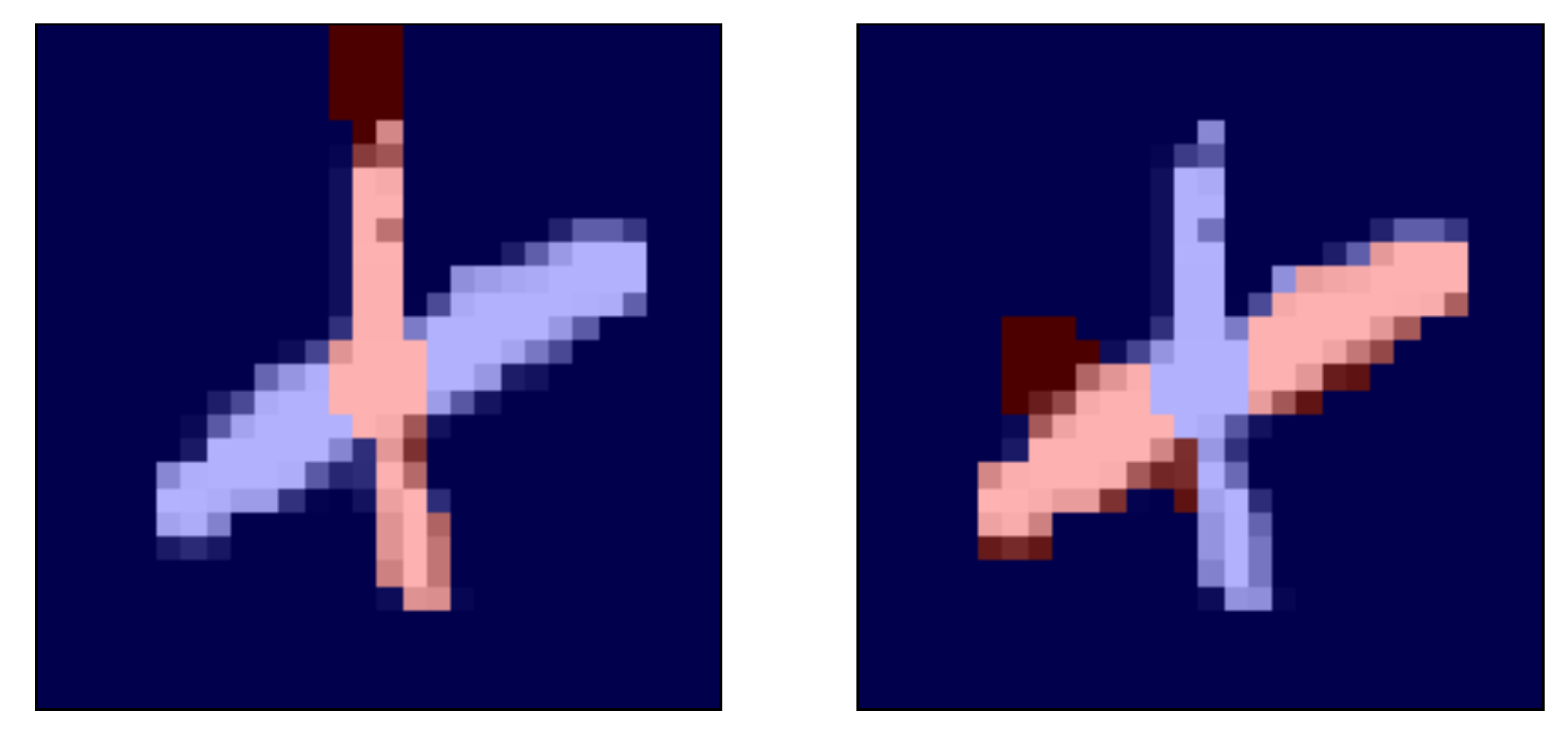}
    \subcaption{Sample from before or after drift.}
    \label{fig:exp:MNISTPlus:+}
    \end{minipage}
    \caption{Explanation for some samples from the MNIST-Plus stream. Images show samples overlayed with LIME relevance  profile for ``before drift'' (left) and ``after drift'' (right). Red indicates high relevance, blue indicates low relevance.}
    \label{fig:exp:MNISTPlus}
\end{figure}
In this experiment, we explore the potential of feature-based explanations for image data. 
\paragraph{Data}
We consider a stream with a single abrupt drift, consisting of vertical lines ($|$), horizontal lines ($-$), and crosses ($+$) which carry characteristics of both types. Before the drift, a vertical line ($|$ or $+$) has to be present, after the drift a horizontal line ($-$ or $+$) has to be present, each class with 50\% rate of occurrence. We obtain the vertical lines as the class ``1'' from the MNIST~\cite{mnist} dataset, the horizontal lines are obtained by rotating those by $90^o$ clockwise, the pluses are then obtained by randomly choosing a horizontal and vertical line and taking the pixel-wise maximum.
\paragraph{Setup}
Our goal is to point out regions in an image, i.e., groups of features, that are particularly relevant for the classification. One suitable explanation approach is the LIME-method. We first use an extremely randomized forest as a model for the drift localization and apply LIME to extract the relevant features.
\paragraph{Results}
Some examples are shown in Figure~\ref{fig:exp:MNISTPlus}. The left image shows the relevances before the drift, while the right image shows those after the drift. Red indicates high relevance and blue low relevance, respectively. As can be seen, pixels that follow the vertical mid-line are strongly associated with ``before drift'' whereas pixels to the left and right, in particular in the central region, are associated with ``after drift''. Thus, the method is capable to capture the main properties of the drift. 

\subsubsection{Counterfactual Explanations for MNIST Streams}
\label{exp:MNIST}
\begin{figure}[!t]
    \centering
    \begin{minipage}{0.99\textwidth}\centering
    \includegraphics[width=0.4\textwidth]{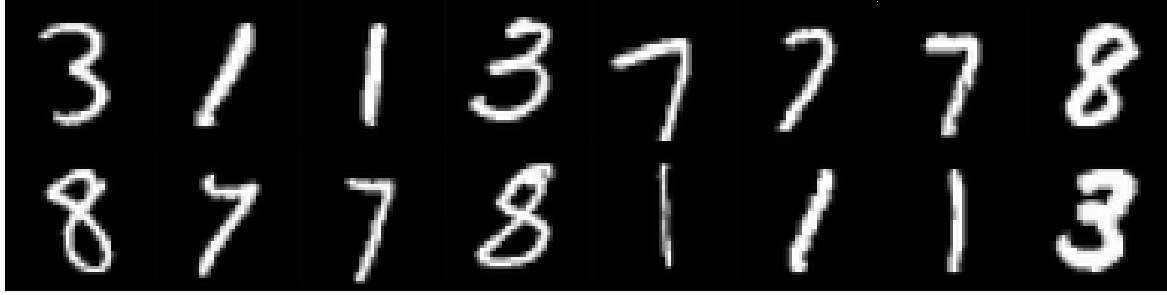}
    \subcaption{Explanation using raw data. The first half of the upper row shows samples before drift, the second half samples after drift respectively. The lower row shows counterfactuals for the samples in the upper row. }
    \label{fig:exp:mnist:raw}%
    \end{minipage}
    \begin{minipage}{0.99\textwidth}
    \includegraphics[width=0.99\textwidth]{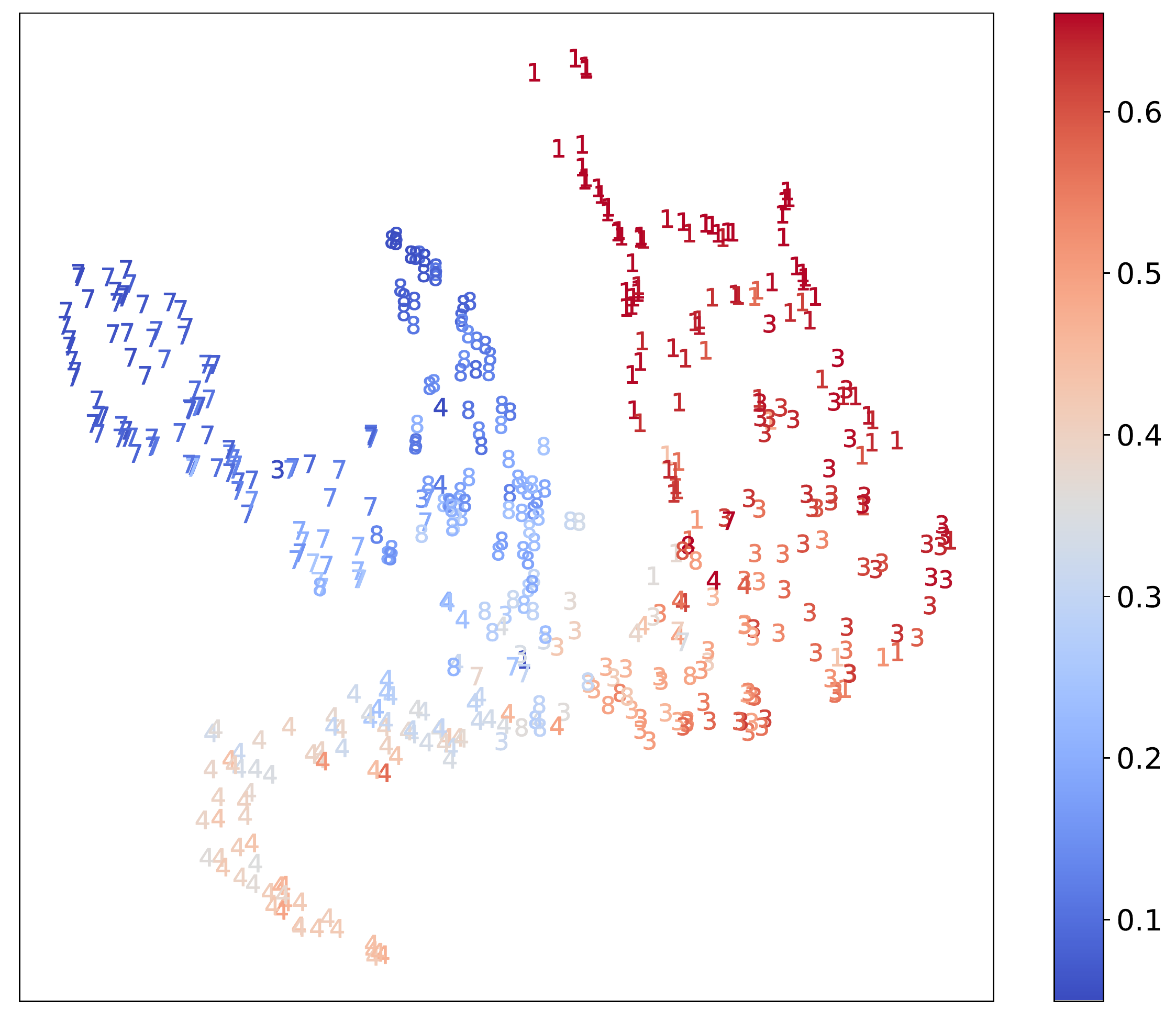}
    \subcaption{Discriminative embedding using t-SNE. Color indicates time-probability ($h(t=1\mid x)$), shape MNIST class.}
    \label{fig:exp:mnist:overview:didi}
    \end{minipage}
    \caption{Visualizations of explanations for MNIST Streams}
    \label{fig:exp:mnist}
\end{figure}

As already pointed out in Section~\ref{sec:explain:combinations}, feature-based explanations are limited in the sense that they do provide insight about which features are relevant for the drift, but not how the drift affected them. One way to tackle this issue is to use counterfactual explanations which are given by a particular similar sample to the one provided except that they belong to a different class, i.e., show a different drifting behavior in our case. A counterfactual so to speak shows the sample as if observed after the drift, providing more information than just the relevant features. In this experiment, we investigate how well this explanation idea translates into practice.

\paragraph{Data}
To showcase the effect of counterfactual explanations we consider a second stream on 
a subset of the $28 \times 28$-pixel black-white MNIST images. The stream has a single, abrupt drift, the digits ``1'', ``3'', and ``4'' are present before and the digits ``7'', ``8'', and ``4'' after the drift, each with the same rate.
Intuitively speaking the drift replaces ``1'' and ``3'' by ``7'' and ``8'' in the stream.

\paragraph{Setup}

We use decision trees to extract the drift information and use
affinity propagation to select the characteristic samples among the drifting ones, as determined by a drift localization~\cite{localization}. 
In order to ensure plausibility, we restricted the set of possible counterfactuals to the training set. 

As in this particular problem, a local explanation by dimensionality reduction is suitable as well, we additionally construct a discriminative embedding using a random forest classifier.

\paragraph{Results}
The counterfactual explanations are presented in Figure~\ref{fig:exp:mnist:raw}. The first half of the upper line presents samples from the data stream before the drift and the second half samples from after the drift. The lower line presents counterfactual explanations of each sample in the upper row, i.e., it shows the user how this sample would have looked like if it had occurred after or before the drift, respectively.

We observe that only the digits ``1'', ``3'', ``7'', ``8'' are considered to be relevant for  the drift.
There is also some tendency to associate the digits ``1'' and ``7'', and ``3'' and ``8''.

The results of the discriminative dimensionality reduction are presented in Figure~\ref{fig:exp:mnist:overview:didi}. As can be seen, the data is separated into three clusters that correlate to the drifting behavior, which overlap mainly at ``3'', ``8'' and ``4'', which can be explained by the optical similarity of ``3'' and ``8'' and the fact that ``4'' is associated with both timepoints. Furthermore, the class probability of the non-drifting samples, i.e., class ``4'', shows a class probability close to 50\% while most samples of the other classes show a strong correlation to the timepoint of observation, as is expected.

As can be seen, the results are very %
promising though only classical methods are applied.
This is not too surprising as MNIST is a rather simple dataset. In the following, we will consider a more complex dataset which can no longer be addressed by classical methods.

\subsubsection{Deep Counterfactuals on ImageNet} 
\label{exp:imagenet}
\begin{figure}[!t]
    \centering
    \begin{minipage}[b]{0.9\textwidth}
    \includegraphics[width=\textwidth]{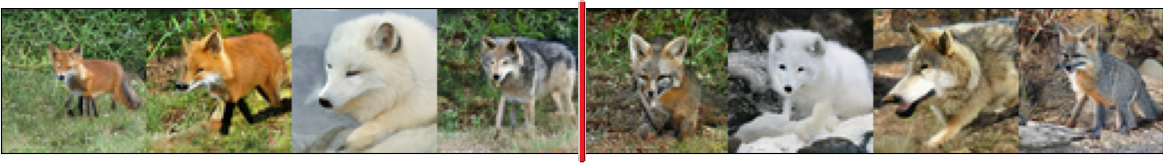}
    \subcaption{Image stream with drift (marked)\label{fig:imagenet-data}}
    \end{minipage}
    \begin{minipage}[b]{0.45\textwidth}
    \includegraphics[width=\textwidth]{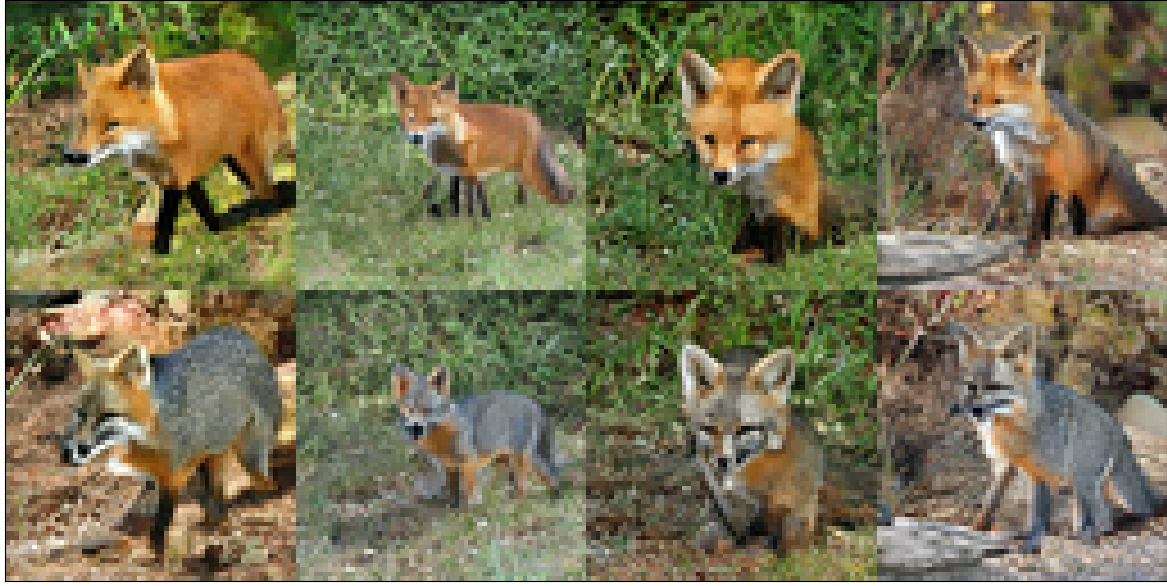}
    \subcaption{Before $\to$ After\label{fig:imagenet-beforeafter}}
    \end{minipage}
    \begin{minipage}[b]{0.45\textwidth}
    \includegraphics[width=\textwidth]{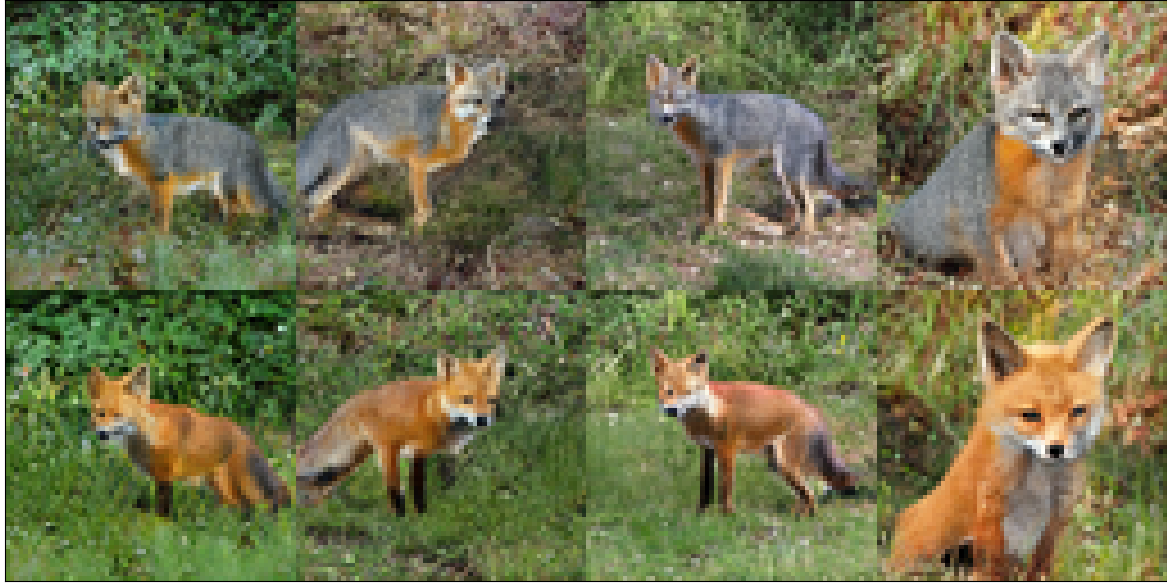}
    \subcaption{After $\to$ Before\label{fig:imagenet-afterbefore}}
    \end{minipage}
    \caption{Illustration of method on ImageNet based data stream. Drift replaces red foxes by gray foxes (a). Explanation (b/c) shows original (Top) and counterfactual (Bottom).}
    \label{fig:Imagenet}
\end{figure}

While we were evaluating our methodology on simple image data in the previous two experiments, we now aim to show its application on more complex datasets. For this purpose, we return to our zoo example which was introduced in Section~\ref{sec:explain:example}. 

\paragraph{Data}
We generate a stream considering images of dogs, cats, hamsters, grey, white, and red wolfs, with red foxes only before, and gray foxes only after the drift. All images are taken from the ImageNet~\cite{imagenet} dataset which consists of $256\times 256$-pixel color photos. Some exemplary samples of the data stream are shown in Figure~\ref{fig:imagenet-data}. The red line indicates the time of the drift.
\paragraph{Setup}
We make use of a pretrained VGG16 network for embedding and a BigGAN~\cite{biggan} for the reconstruction. We perform the computations of the counterfactuals in the latent space using decision trees as a model for drift localization and apply $k$-means to the drifting samples to obtain the characteristic samples. The counterfactuals are computed using CEML~\cite{ceml}.
\paragraph{Results}
The obtained results are shown in Figures~\ref{fig:imagenet-beforeafter}, \ref{fig:imagenet-afterbefore}. Again, the upper rows present the observed samples while the lower rows show the counterfactual explanations of the corresponding samples in the upper row. As can be seen, our method correctly identifies red and gray foxes as the drift-inducing feature, which is then exchanged in the production of the counterfactuals. Notice that the main feature changed by this procedure is the fur color whereas, for example, the posture of the fox is nearly unchanged. However, we made use of a version of BigGAN that respects the image class and thus captured the drift by means of counterfactuals particularly well. It is thus questionable whether or not this approach also works comparably well if we do not consider drift that aligns well with what the deep model is designed to process.

\section{Discussion and Future Work}
\label{sec:conclusion}
In this work, we considered the problem of explaining concept drift by means of model-based explanations that provide insight into the characteristics of the drift. Our approach is model and explanation independent and 
we demonstrated its behavior in several examples. The empirical results demonstrate that this proposal constitutes a promising approach as regards drift explanation in an intuitive fashion. The technology is not limited to a finite amount of timepoints nor does it require drift detection in order to be applicable. 
Although the methodology can be broadly justified theoretically, understanding feature-wise drift behavior is still an unsolved problem which requires a deeper theoretical understanding. Considering the large variety of graphical models describing real-world data, the proposed data generation in Section~\ref{exp:bayesnet} where we used a Bayesian network to create the dataset, could give rise to further theoretical and methodological research in the area of structured analysis of drifting features. 

Moreover, applying the technology to complex data such as images requires appropriate preprocessing, e.g., by a deep neural network, which may render some of the explanatory methods useless.
Future work could try to deal with this problem by taking ideas from transfer learning into account. Since deep convolutional networks tend to learn a universal representation of the data they might be useful as a universal preprocessing for the data at hand.
Besides, examining our proposed methodology in additional important real-world tasks, as for example in anomaly detection  and explainable online learning models might be a promising research direction.

\section*{Acknowledgement}
Funding in the frame of the BMBF project TiM, 05M20PBA and the BMWi project KI-Marktplatz, 01MK20007E is gratefully acknowledged.

\bibliographystyle{elsarticle-num}
\bibliography{bib}

\end{document}